\documentclass{article}

\usepackage[preprint]{neurips_2025}

\usepackage{amsmath,amsfonts,bm}

\def\eqref#1{equation~\ref{#1}}

\def\1{\bm{1}}

\DeclareMathAlphabet{\mathsfit}{\encodingdefault}{\sfdefault}{m}{sl}
\SetMathAlphabet{\mathsfit}{bold}{\encodingdefault}{\sfdefault}{bx}{n}

\def\gC{{\mathcal{C}}}

\def\gG{{\mathcal{G}}}

\def\gR{{\mathcal{R}}}

\newcommand{\E}{\mathbb{E}}

\usepackage[utf8]{inputenc}
\usepackage[T1]{fontenc}
\usepackage{url}
\usepackage{booktabs}
\usepackage{amsmath}
\usepackage{amssymb}
\usepackage{amsfonts}
\usepackage{nicefrac}
\usepackage{microtype}
\usepackage{xcolor}
\usepackage{placeins}
\usepackage{graphicx}
\usepackage{subcaption}
\usepackage{caption}
\usepackage{enumitem}
\usepackage{multirow}
\usepackage{algorithm}
\usepackage[noend]{algpseudocode}
\usepackage[most]{tcolorbox}

\usepackage{bbm}

\definecolor{mydarkblue}{rgb}{0,0.08,0.45}
\usepackage[colorlinks,citecolor=mydarkblue,urlcolor=mydarkblue,linkcolor=mydarkblue]{hyperref}
\usepackage[capitalize,noabbrev]{cleveref}

\usepackage{xspace}
\newcommand{\scorpio}{\textsc{Scorpio}\xspace}

\makeatletter
\renewcommand\Affilfont{\normalfont\sffamily\bfseries\fontsize{8}{10}\selectfont}
\makeatother

\title{On the Limits of Test-Time Compute for Self-Evaluation}

\author{%
  Anirudh Buvanesh$^{1,2}$,\,\,
  Khaled Rouissi$^{1,2}$,\,\,
  Sujin Yun$^{1,2}$,\,\,
  Johan Obando-Ceron$^{1,2}$,\,\,
  Morgane M Moss$^{3}$,\\
  Samuel Lavoie$^{4}$,\,\,
  Michael Noukhovitch$^{1,2}$,\,\,
  Laurent Charlin$^{\dagger\,1,5,6}$,\,\,
  Aaron Courville$^{\dagger\,1,2,6}$\\
  {\normalfont\fontsize{8}{10}\selectfont
  $^1$Mila \ \ $^2$Universit\'e de Montr\'eal \ \ $^3$Nous Research \ \ $^4$Google \ \ $^5$HEC Montr\'eal \ \ $^6$Canada CIFAR AI Chair}\\
  {\normalfont\scriptsize $^\dagger$Equal Advising}
}

\definecolor{promptbg}{RGB}{245, 248, 250}
\definecolor{promptborder}{RGB}{76, 175, 147}
\newtcolorbox{promptbox}[1]{%
  colback=promptbg, colframe=promptborder,
  title={\textbf{#1}}, coltitle=white, fonttitle=\small\bfseries,
  breakable, enhanced,
  fontupper=\normalfont,
}

\definecolor{greyboxbg}{RGB}{238, 238, 238}
\definecolor{greyboxframe}{RGB}{190, 190, 190}
\newtcolorbox{greybox}[1]{%
  colback=greyboxbg, colframe=greyboxframe,
  coltitle=black, fonttitle=\small\bfseries,
  title={#1},
  boxrule=0.4pt, arc=1pt,
  breakable, enhanced,
}

\begin{document}

\maketitle
\suppressfloats[t]

\begin{abstract}
Self-verification and aggregation have emerged as popular approaches for bridging the gap between $\mathrm{pass@1}$ and $\mathrm{pass@}N$. However, increasing the number of candidates does not consistently narrow the gap between final-answer accuracy and $\mathrm{pass@}N$. To understand this limitation, we use ranking error to measure how reliably models distinguish their correct generations from incorrect ones. We show that ranking error admits a bias-variance decomposition, with both components shaped by algorithmic design choices and inference compute. Across our experiments, additional compute reduces variance, while bias initially decreases before approaching a plateau.  Guided by this analysis, we introduce \scorpio, a \textit{simple} and \textit{general} method that jointly scores randomly grouped candidates. Across math, code, and puzzle benchmarks, \scorpio improves verification, tree search, and aggregation. For verification, \scorpio improves accuracy over the strongest baseline by up to $7$ percentage points while matching or exceeding that baseline’s peak accuracy with $17$--$42\%$ lower inference cost. For tree search, using \scorpio to score partial solutions and select the final answer improves accuracy by up to $10$ percentage points over prior methods. Finally,  \scorpio improves the scaling of aggregation-based methods by  $11$ percentage points on coding tasks.

\medskip
\noindent{\small\sffamily\bfseries Code Repository:} {\small\url{https://github.com/anirudhb11/scorpio}}\\[3pt]
{\small\sffamily\bfseries Correspondence:} {\small\texttt{anirudh.buvanesh@mila.quebec}}
\end{abstract}

\section{Introduction}
\label{sec:introduction}

Scaling test-time compute has enabled substantial progress in the reasoning capabilities of large language models (LLMs)~\citep{snell2024scaling,openai2026astra,anthropic2026claude51}. By spending more compute on a problem, models can explore alternative reasoning strategies~\citep{wang2022self}, verify intermediate results~\citep{yao2023tot}, and revise earlier attempts before returning an answer~\citep{madaan2023selfrefine}. These capabilities have enabled models to achieve gold-level performance in math olympiads~\citep{luong2025geminiimo, wei2025imoproofs}, complete long-horizon agentic tasks~\citep{kwa2025measuring}, and, more recently, propose a solution to an open Millennium Prize Problem~\citep{openai2026navierstokesblog}.

\begin{figure}[t]
  \centering
  \makebox[\textwidth]{\hspace{0.0153\textwidth}\includegraphics[width=0.984\textwidth]{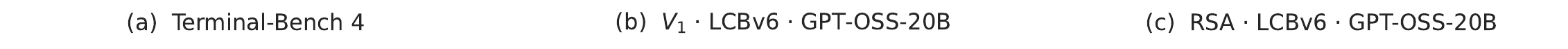}\hspace{-0.0153\textwidth}}\\[-0.35em]
  \makebox[\textwidth]{\hspace{0.0153\textwidth}\includegraphics[width=0.984\textwidth]{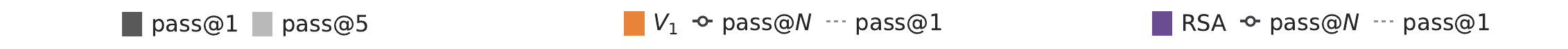}\hspace{-0.0153\textwidth}}\\[-0.2em]
  \begin{subfigure}[t]{0.325\textwidth}
    \includegraphics[width=\textwidth]{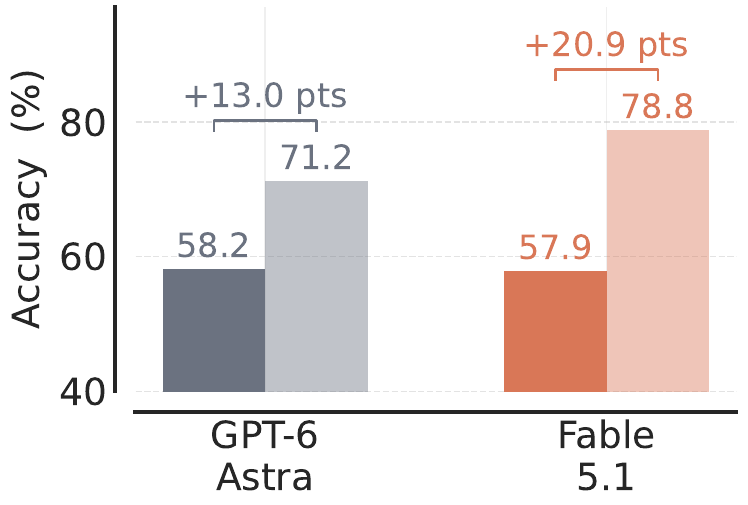}
    \phantomsubcaption\label{fig:intro_a}
  \end{subfigure}\hfill
  \begin{subfigure}[t]{0.325\textwidth}
    \includegraphics[width=\textwidth]{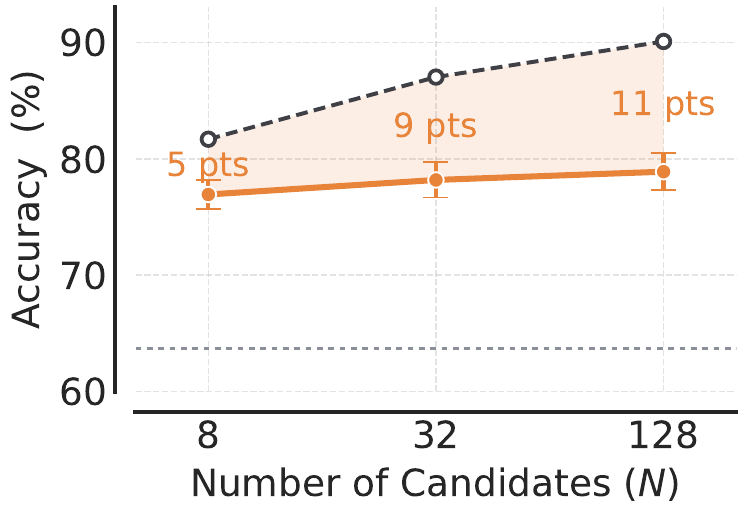}
    \phantomsubcaption\label{fig:intro_b}
  \end{subfigure}\hfill
  \begin{subfigure}[t]{0.325\textwidth}
    \includegraphics[width=\textwidth]{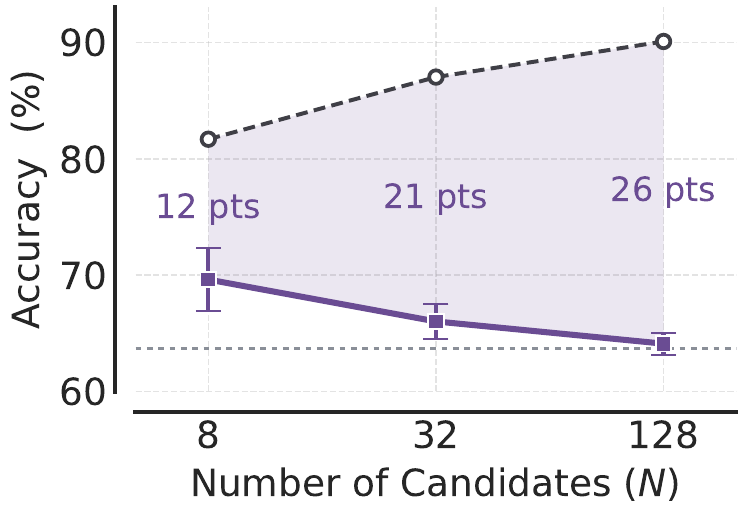}
    \phantomsubcaption\label{fig:intro_c}
  \end{subfigure}
  \caption{\textbf{Sampling more candidates improves the chance of a correct answer, but selecting it does not scale well with $N$.} \textbf{(a)} The gap between $\mathrm{pass@}1$ and $\mathrm{pass@}N$ exists even for frontier models on agentic tasks.\protect\footnotemark{} \textbf{(b)} As $N$ increases, the gap between $\mathrm{pass@N}$ and the selection accuracy of V$_1$~\citep{singh2026v_1}, a recent self-verification method for scaling test-time compute, widens. \textbf{(c)} A similar trend holds for Recursive Self-Aggregation (RSA) ~\citep{venkatraman2025rsa}. For (b) and (c), we run each method until accuracy plateaus; refer \autoref{app:selection-passn-gap} for further details.}
  \label{fig:intro}
\end{figure}

A common building block of these approaches is sampling multiple candidate solutions~\citep{gu2026understanding}, which increases the chance of obtaining at least one correct answer compared to a single attempt. The resulting gap between the accuracy of a single attempt and the probability that at least one of $N$ attempts succeeds, the $\mathrm{pass@1}$ versus $\mathrm{pass@}N$ gap, remains significant even for frontier models on agentic tasks. On Terminal-Bench 4, GPT-6 Astra gains 13.0 points and Fable 5.1 gains 20.9 points when moving from $\mathrm{pass@1}$ to $\mathrm{pass@5}$ (\autoref{fig:intro_a}). Several approaches use additional compute to bridge this gap, ranging from verifying candidates to identify a correct one~\citep{cobbe2021training,singh2026v_1,kwok2026llmasaverifiergeneralpurposeverificationframework} to iteratively synthesizing better solutions from a pool of existing candidates~\citep{venkatraman2025rsa, madaan2025rethinking}.
However, the ability of these methods to \textit{select} a good candidate does not consistently scale with the number of candidates. As shown in \autoref{fig:intro_b} and \autoref{fig:intro_c}, the gap between selection accuracy and oracle accuracy widens with the number of candidates, while aggregation-based methods show a decline in accuracy.

We investigate this issue by adapting the pairwise ranking loss from the ranking literature~\citep{shivaswamy2021bias} to measure how accurately an LLM ranks its own solutions. We show that this loss admits a bias-variance decomposition, where bias captures systematic errors in pairwise judgments and variance captures fluctuations across repeated evaluations. This decomposition allows us to study how design choices in test-time scaling algorithms, such as the number of candidates evaluated in a single LLM call and how they are grouped, affect ranking quality. We find that variance generally decreases with inference compute, while bias appears to plateau within the evaluated budgets.  This residual bias captures errors that persist even at high inference budgets, making bias reduction a key step toward closing the gap between selection accuracy and oracle performance.

Guided by this analysis, we introduce \underline{S}coring \underline{C}andidates \underline{O}ver \underline{R}andom \underline{P}artit\underline{io}ns (\scorpio), a simple scoring strategy that repeatedly partitions candidates into random groups and scores them jointly. Despite its simplicity, this design reduces variance faster than prior scoring methods while keeping bias comparable, yielding similar or lower ranking error. Across math, code, and puzzle benchmarks, \scorpio improves peak verification accuracy over the strongest baseline by up to \textbf{7} percentage points, while matching or exceeding that baseline's peak accuracy with \textbf{17--42\%} lower total inference cost. Beyond verification, \scorpio can also be applied to other test-time scaling methods. For tree search, it effectively identifies promising partial solutions to continue, improving over the strongest baseline by \textbf{10} percentage points on LiveCodeBench~v6. Further, \scorpio improves the test-time scaling of aggregation-based methods, reversing the decline in accuracy with larger candidate pools highlighted in \autoref{fig:intro_c}. Despite these improvements, the residual bias identified above persists even under \scorpio, leaving a gap between selection accuracy and $\mathrm{pass@}N$ at high inference budgets.

\footnotetext{Accuracies from the \href{https://hub.harborframework.com/datasets/terminal-bench/terminal-bench/4?tab=leaderboard&leaderboard=4-0-0}{Terminal-Bench 4 leaderboard}.}

\section{Related Work}
\label{sec:related_work}
\textbf{LLMs as judges.}
LLMs are widely used to judge the quality of model-generated responses~\citep{zheng2023judging}. More recently, these judgments have provided a promising axis for scaling test-time compute on reasoning tasks, where repeated evaluations have been shown to improve the ability to select a correct solution from a set of generated candidates~\citep{singh2026v_1,kim2026rtv,kwok2026llmasaverifiergeneralpurposeverificationframework}. However, prior work has also documented limitations of LLM judges, including preferences for their own outputs~\citep{panickssery2404llm,yang2026quantifying}, sensitivity to candidate order~\citep{panickssery2404llm}, and inconsistencies across repeated judgments~\citep{haldar-hockenmaier-2025-rating,yagubyan2026coin,norman2026reliability}. Our work examines both the potential and the limitations of self-verification through a bias--variance decomposition of ranking error. This decomposition offers a unified account of these limitations: inconsistency across repeated judgments corresponds to the variance term, while systematic preferences such as self-preference and order sensitivity plausibly manifest as bias that additional compute cannot remove. We find that a residual bias remains as inference compute increases, while variance decreases monotonically. While prior work has discussed variance reduction~\citep{kwok2026llmasaverifiergeneralpurposeverificationframework}, our work provides a more nuanced view by highlighting that the rate of variance reduction depends on algorithmic design choices. For instance, we observe slower variance reduction with inference compute when pairing similar candidates, as in $\mathrm{V}_1$~\citep{singh2026v_1}, than when pairing candidates randomly.

\textbf{Bias-variance decomposition of losses.} Bias-variance decompositions are a standard tool for analyzing the tradeoffs involved in training machine learning models~\citep{geman1992neural}. Such decompositions have been developed for squared loss~\citep{bishop2006pattern}, zero-one loss~\citep{kohavi1996zerooneloss,domingos2000unified}, and ranking objectives~\citep{shivaswamy2021bias}. We extend this perspective to test-time scaling by studying how bias and variance in rankings evolve as a function of inference compute. In this setting, stochasticity arises from temperature-based sampling by an LLM judge, rather than from training predictors on sampled training data. Finally, we use this decomposition to motivate simpler design choices in recent test-time algorithms, showing that random pairing outperforms tournament-based pairing at equal compute. 
\autoref{app:ext-rel-work} provides a more nuanced distinction from algorithmically similar approaches in the retrieval literature.

\section{Background}
\label{sec:background}

We consider three settings for scaling test-time compute: selecting among complete solutions, guiding search over partial solutions, and aggregating multiple solutions to generate new ones. Given a problem $x$, a language model $\pi_\theta$ generates a solution $y_{1:T}\sim\pi_\theta(\cdot\mid x)$, where $T$ denotes its length in tokens. Sampling $N$ such solutions produces a candidate pool $\gC=\{y^{(1)},\ldots,y^{(N)}\}$.

\textbf{Candidate scoring.}
A scoring procedure $f$ assigns a score to each candidate in $\gC$. We focus on self-evaluation, where the generating model $\pi_\theta$ judges its own responses using a prompt $P_{\text{judge}}$. Candidates can be evaluated independently through pointwise scoring~\citep{weng2023large}, yielding $s^{(i)}\sim\pi_\theta(\cdot\mid P_{\text{judge}},x,y^{(i)})$, or jointly within a shared context~\citep{singh2026v_1}, yielding $(s^{(1)},\ldots,s^{(N)})\sim\pi_\theta(\cdot\mid P_{\text{judge}},x,\gC)$. In either case, more inference compute can be spent by repeating evaluations and combining the resulting scores~\citep{kwok2026llmasaverifiergeneralpurposeverificationframework}. For instance, in pointwise scoring, the final score can be computed as $\hat{s}^{(i)}=\frac{1}{M}\sum_{m=1}^{M}s_m^{(i)}$, where $s_m^{(i)}$ is the score from the $m$-th evaluation. This requires $M \cdot N$ LLM calls to evaluate all $N$ candidates $M$ times. The highest-scoring candidate is then selected, with $i^*=\arg\max_i\hat{s}^{(i)}$.

\textbf{Candidate search.} 
The same scoring procedure can be used during generation to evaluate partial solutions $y_{1:t}$ ($t < T$). Search proceeds in stages: at each stage, the retained partial solutions are extended, scored, and pruned. Beam search~\citep{Lowerre1976TheHS} retains a fixed number of top-scoring candidates, whereas we retain the top-scoring fraction $\rho$ of them, as described in \autoref{sec:tree_search}.

\textbf{Candidate aggregation.}
Aggregation generates new solutions by conditioning on existing candidates. Given a subset $\mathcal{S}\subseteq\gC$ and a prompt $P_{\text{agg}}$ instructing the model to combine and improve these solutions, a new solution is sampled as $y'\sim\pi_\theta(\cdot\mid P_{\text{agg}},x,\mathcal{S})$. Repeating this operation over $N$ randomly sampled subsets produces a new pool of $N$ candidates, which can be used in subsequent aggregation rounds~\citep{venkatraman2025rsa,madaan2025rethinking}.

Across these settings, effective test-time scaling requires accurately ranking candidates. In selection, a ranking error affects only the final choice. In search and aggregation, errors compound: pruning a promising partial solution or retaining a poor candidate for aggregation affects all subsequent generations.
In the next section, we define the pairwise ranking loss to measure ranking quality and analyze how its bias and variance change with inference compute.

\section{Decomposing Ranking Error Into Bias and Variance}
\label{sec:measuring_ranking}

We measure how accurately a language model ranks its own generations by examining whether correct solutions receive higher scores than incorrect ones. Let $\gC^+$ and $\gC^-$ denote the subsets of correct and incorrect solutions in $\gC$, such that $\gC = \gC^+ \cup \gC^-$. Following \citet{cortes2003aucvserror}, we define the pairwise ranking error of the scoring procedure $f$ as:

\begin{align}
\gR(f)
=
\begin{cases}
\displaystyle
\frac{1}{|\gC^+|\,|\gC^-|}
\sum_{y^{(i)}\in\gC^-}
\sum_{y^{(j)}\in\gC^+}
\mathbbm{1}\!\left[
\hat{s}^{(i)} \geq \hat{s}^{(j)}
\right],
& \text{if } |\gC^+|>0 \text{ and } |\gC^-|>0, \\[6pt]
0,
& \text{otherwise}.
\end{cases}
\label{eqn:ranking-error}
\end{align}

In other words, the ranking error is the fraction of correct--incorrect pairs for which $f$ assigns a higher score to the incorrect solution. This error is stochastic because the scores used to rank candidates are outputs of temperature-based sampling from the LLM. Evaluating candidates jointly can introduce further stochasticity when pairings are randomized, since the same candidate may receive different scores depending on its opponent. We therefore analyze the expected ranking error $\E[\gR(f)]$, where the expectation is over randomness in the scoring procedure. For each correct--incorrect pair, we define
$Z_{ij}=\mathbbm{1}[\hat{s}^{(i)} \geq \hat{s}^{(j)}]$.
Since $Z_{ij}$ is binary, $Z_{ij}^2=Z_{ij}$.
Adding and subtracting $(\E[Z_{ij}])^2$ gives
\begin{align}
\E[Z_{ij}]
&= 
   \E[Z_{ij}^2] - \bigl(\E[Z_{ij}]\bigr)^2
   + \bigl(\E[Z_{ij}]\bigr)^2 = \operatorname{Var}(Z_{ij})
   + \bigl(\E[Z_{ij}]\bigr)^2.
\end{align}
Averaging over all correct--incorrect pairs, we obtain
\begin{align}
\E[\gR(f)]
=
\frac{1}{|\gC^+|\,|\gC^-|}
\sum_{y^{(i)}\in\gC^-}
\sum_{y^{(j)}\in\gC^+}
\left[
\underbrace{\bigl(\E[Z_{ij}]\bigr)^2}_{\text{Squared bias}}
+
\underbrace{\operatorname{Var}(Z_{ij})}_{\text{Variance}}
\right],
\label{eqn:bias-variance}
\end{align}

Recall that $Z_{ij}=1$ when the incorrect solution is scored higher than the correct one. Its expectation measures how often this error occurs, so $(\E[Z_{ij}]-0)^2$ captures the squared bias relative to the desired value of zero. Its variance measures how consistently the scoring procedure orders the pair, regardless of whether the ordering is correct.
Writing $p_{ij}=\E[Z_{ij}]$ for the misranking probability of a pair, the two terms are $p_{ij}^2$ and $p_{ij}(1-p_{ij})$. Although both are determined by $p_{ij}$, they respond differently to changes in it: variance vanishes as $p_{ij}$ approaches either $0$ or $1$, whereas squared bias vanishes only as $p_{ij}$ approaches $0$. Averaged over pairs, a decrease in variance without a corresponding decrease in bias therefore indicates that the scoring procedure orders pairs more consistently, but that some pairs are consistently misranked.

\textbf{Estimating bias and variance.}
At a given inference budget, we repeat the scoring procedure $f$ $R$ times to obtain $R$ scores for each candidate. These scores yield $R$ rankings over $N$ candidates. We then estimate the misranking probability for each correct-incorrect pair as $\hat{p}_{ij}=\frac{1}{R}\sum_{r=1}^{R}Z_{ij}^{(r)}$, where $Z_{ij}^{(r)}=1$ when the incorrect solution is scored higher than the correct one in the $r$-th ranking. The pair's squared bias and variance are estimated with the unbiased estimators $\hat{p}_{ij}^{2}-\frac{\hat{p}_{ij}(1-\hat{p}_{ij})}{R-1}$ and $\frac{R}{R-1}\,\hat{p}_{ij}(1-\hat{p}_{ij})$, which sum to $\hat{p}_{ij}$. To reduce the variance of these estimators, we use $R=20$ repetitions throughout our experiments. Averaging these estimates over all correct-incorrect pairs gives estimates of the squared bias and variance terms in \autoref{eqn:bias-variance}. \autoref{app:bias-var-computation} provides further details on how bias and variance are estimated.

\begin{algorithm}[t]
\caption{\scorpio Selection}
\label{alg:scorpio}
\begin{algorithmic}[1]
\State \textbf{Inputs:} problem $x$; pool $\gC=\{y^{(1)},\ldots,y^{(N)}\}$; policy $\pi_\theta$;
group size $K$; repeats $M$; groupwise scorer $f$.
\State $\hat s \gets [\,]$ \Comment{candidate scores}
\State \textbf{Score Candidates:}
\For{$m \gets 1$ to $M$}
    \State $\sigma_m \gets$ random permutation of $[N]$ \Comment{independent partition per repeat}
    \For{$j \gets 1$ to $N/K$}
        \State $\gG \gets \sigma_m[(j-1)K+1:jK]$\Comment{Take the next $K$ candidates indices in the shuffled order}
        \State $\big(r^{(i)}_m\big)_{i\in\gG} \gets f\big(x,(y^{(i)})_{i\in\gG}\big)$ \Comment{$K$ candidates scored together}
    \EndFor
\EndFor
\State \textbf{Select Candidate:}
\State $\hat s^{(i)} \gets \frac{1}{M}\sum_{m=1}^{M} r^{(i)}_m$ \Comment{average each candidate's $M$ ratings}
\State $i^\ast \gets \arg\max_{i\in[N]} \hat s^{(i)}$ 
\State \textbf{return} $y^{(i^\ast)}$
\end{algorithmic}
\end{algorithm}

\section{\texorpdfstring{\scorpio: \underline{S}coring \underline{C}andidates \underline{O}ver \underline{R}andom \underline{P}artit\underline{io}ns}{SCORPIO: Scoring Candidates Over Random Partitions}}

While previous methods have focused on evaluating candidates individually or in pairwise comparisons, we argue that contrasting a larger group inside a single LLM call can reduce bias and variance. We propose \scorpio (\autoref{alg:scorpio}), a simple instantiation of the scoring procedure $f$ that randomly partitions a pool $\gC$ of $N$ candidates into groups of size $K$ and \textit{jointly} scores the candidates in each group in a single call. We run $M$ rounds of scoring, creating fresh random partitions each round and using $MN/K$ LLM calls in total. 

\begin{figure}[t]
  \centering
  \includegraphics[width=\textwidth]{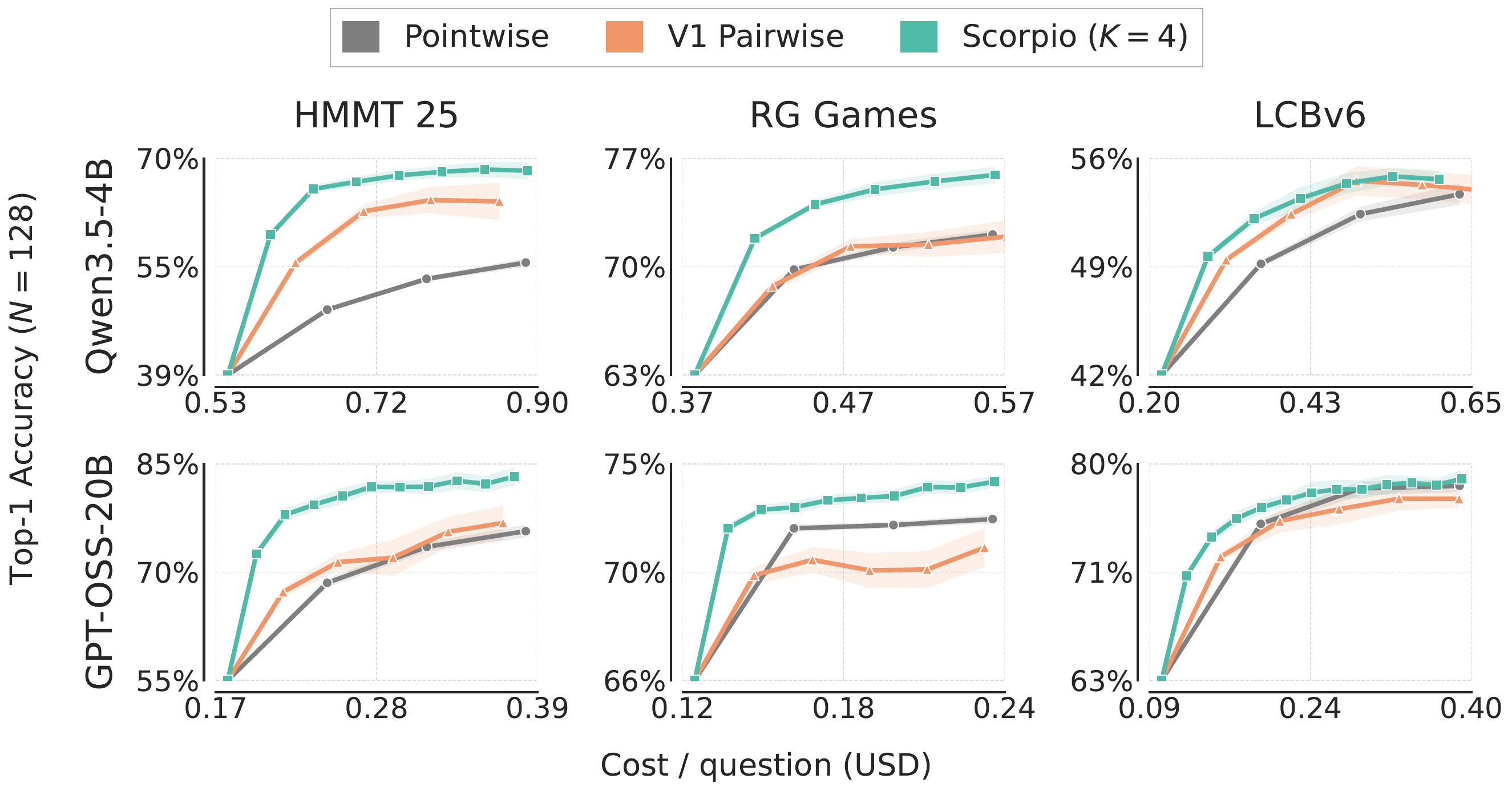}
  \caption{\textbf{\scorpio scales better with inference compute than prior verification methods across math (HMMT), puzzle (RG Games), and code benchmarks (LCBv6).} We compare \scorpio's scoring with V1~\citep{singh2026v_1}, a recent test-time scaling algorithm that scores candidates through pairwise comparisons between similar candidates, and pointwise scoring, where LLMs evaluate candidates in isolation. \scorpio jointly scores candidates in groups of four. Inference compute is measured as the total dollar cost of all input and output tokens, and is scaled by repeating candidate evaluations. Results are averaged over 20 seeds, with shaded regions indicating $95\%$ confidence intervals.}
  \label{fig:wbest_methods}
\end{figure}

\textbf{Setup.}
We evaluate Qwen3.5-4B~\citep{qwen2026qwen35} in instruct mode and GPT-OSS-20B~\citep{agarwal2025gpt} at medium reasoning effort on math with HMMT Feb 2025, code with LiveCodeBench~\citep{jain2025livecodebench}, and general reasoning with Reasoning Gym games~\citep{stojanovski2026reasoning}. Following \citet{hubotter2026reinforcement}, we consider the LCBv6 subset which contains $131$ questions released between February and May 2025. For Reasoning Gym games, we use the $100$-problem split from \citet{venkatraman2025rsa}. Throughout our experiments, test-time compute is measured by the total dollar cost of all LLM calls. We use input and output prices of \$0.03 and \$0.10 per million tokens for GPT-OSS-20B, and \$0.20 per million tokens for both input and output for Qwen3.5-4B.\footnote{Prices obtained from \url{https://pricepertoken.com} as of September 13, 2026.} \autoref{app:exp-setup} describes the sampling parameters and cost calculation for each test-time scaling method, and \autoref{app:prompts} lists the evaluation prompts.

\begin{figure}[t]
  \centering
  \includegraphics[width=\textwidth]{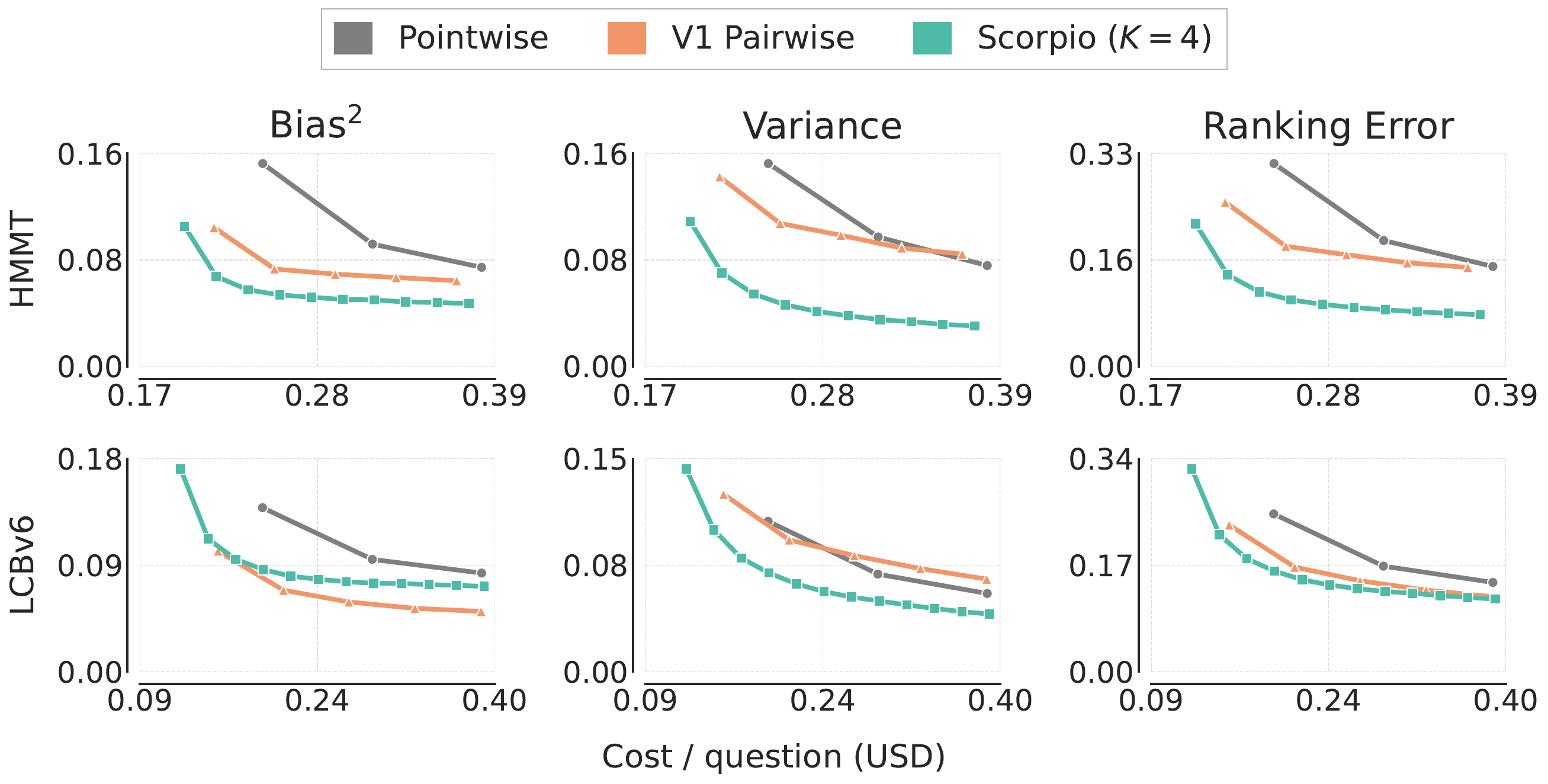}
  \caption{\textbf{Decomposition of ranking error into bias and variance.} Results for GPT-OSS-20B on HMMT (math) and LCBv6 (code). As inference compute increases, variance generally decreases, while bias decreases  before approaching a plateau (see \autoref{app:bias-dec} for why bias can decrease with inference compute). Despite its simpler design, \scorpio reduces variance faster than pointwise scoring and V1's~\citep{singh2026v_1} pairwise scoring, while achieving similar or lower ranking error. Similar trends are observed for Qwen3.5-4B (see \autoref{fig:qwen_biasvar}).}
  \label{fig:main_biasvar}
\end{figure}

\subsection{Test-Time Scaling for Verification}
\label{sec:verification}

In self-verification, the goal is to select the single best candidate from a pool of $N=128$ candidates generated by the model. The ultimate goal is to narrow the gap between $\mathrm{pass@}1$ and $\mathrm{pass@}N$. We compare against the standard pointwise approach, which generates a score for each sample $s^{(1)} \sim f(y^{(1)})$ and a recent state-of-the-art method, V1 \citep{singh2026v_1} which combines pairwise comparisons $s^{(1)},s^{(2)} \sim f(y^{(1)},y^{(2)})$ with a tournament-style partitioning, iteratively comparing better-scoring candidates with each other. Following \citet{singh2026v_1}, we cap the verification budget at the dollar cost of three rounds of pointwise scoring.

\textbf{\scorpio scales better than pointwise and tournament-based pairwise scoring.}
\autoref{fig:wbest_methods} compares \scorpio with group size $K=4$ against pointwise scoring and $\mathrm{V}_1$. Across HMMT and Reasoning Gym games, \scorpio improves peak accuracy over the strongest baseline by \textbf{2--7} percentage points and matches or exceeds that baseline's peak accuracy with \textbf{17--41\%} lower inference cost. On LCBv6, all methods attain similar peak accuracy. With GPT-OSS-20B, \scorpio matches the strongest baseline's peak accuracy with \textbf{18\%} lower inference cost. We attribute these gains to (1) LLM's ability to better compare candidates \textit{within} a context than between separate LLM calls and (2) more efficient use of LLM context and calls by combining multiple inputs.  \scorpio also compares favorably with Probabilistic Pivot Tournament (PPT)~\citep{kwok2026llmasaverifiergeneralpurposeverificationframework} (see \autoref{fig:app_ppt_vs_scorpio}).

\textbf{Variance reduces with compute, but ranking bias plateaus.}
Despite the improvements provided by \scorpio, its selection accuracy plateaus well below $\mathrm{pass@}128$, leaving an average gap of \textbf{14.3} percentage points across the six settings in \autoref{fig:wbest_methods}. To better understand why additional verification compute does not close this gap, we revisit the ranking error and its bias-variance decomposition introduced in \autoref{sec:measuring_ranking}. \autoref{fig:main_biasvar} compares \scorpio, pointwise scoring, and $\mathrm{V}_1$ across the two datasets using GPT-OSS-20B. Across the evaluated methods, variance generally decreases as inference compute increases, while bias typically decreases initially before approaching a plateau.\footnote{See \autoref{app:bias-dec} for a discussion of how ranking bias can decrease with inference compute.} In other words, additional compute makes rankings more stable, but incorrect solutions still receive higher scores than correct ones. Extrapolating the variance trend, we estimate that bias accounts for $91\%$ to $99\%$ of the ranking error as compute grows across both models and all three datasets (\autoref{app:var-rate}). 

\textbf{Random grouping reduces variance faster, while tournament pairing can lower bias.}
Although \scorpio achieves similar or lower ranking error than $\mathrm{V}_1$ (\autoref{fig:main_biasvar}), the two methods differ in how this error splits into bias and variance. $\mathrm{V}_1$ pairs candidates with similar scores and consistently shows higher variance than \scorpio, likely because such comparisons are harder than random ones and yield less consistent rankings across repeated evaluations. On the other hand, this pairing strategy can lower bias: on LCBv6, $\mathrm{V}_1$ attains lower bias than \scorpio. To isolate the effect of the pairing strategy, we compare $\mathrm{V}_1$ with \scorpio at $K=2$ (\autoref{fig:app_v1_vs_k2_biasvar}) and observe similar trends, showing that \scorpio's faster variance reduction is not solely due to its larger group size.

\begin{figure}[t]
  \centering
  \includegraphics[width=\textwidth]{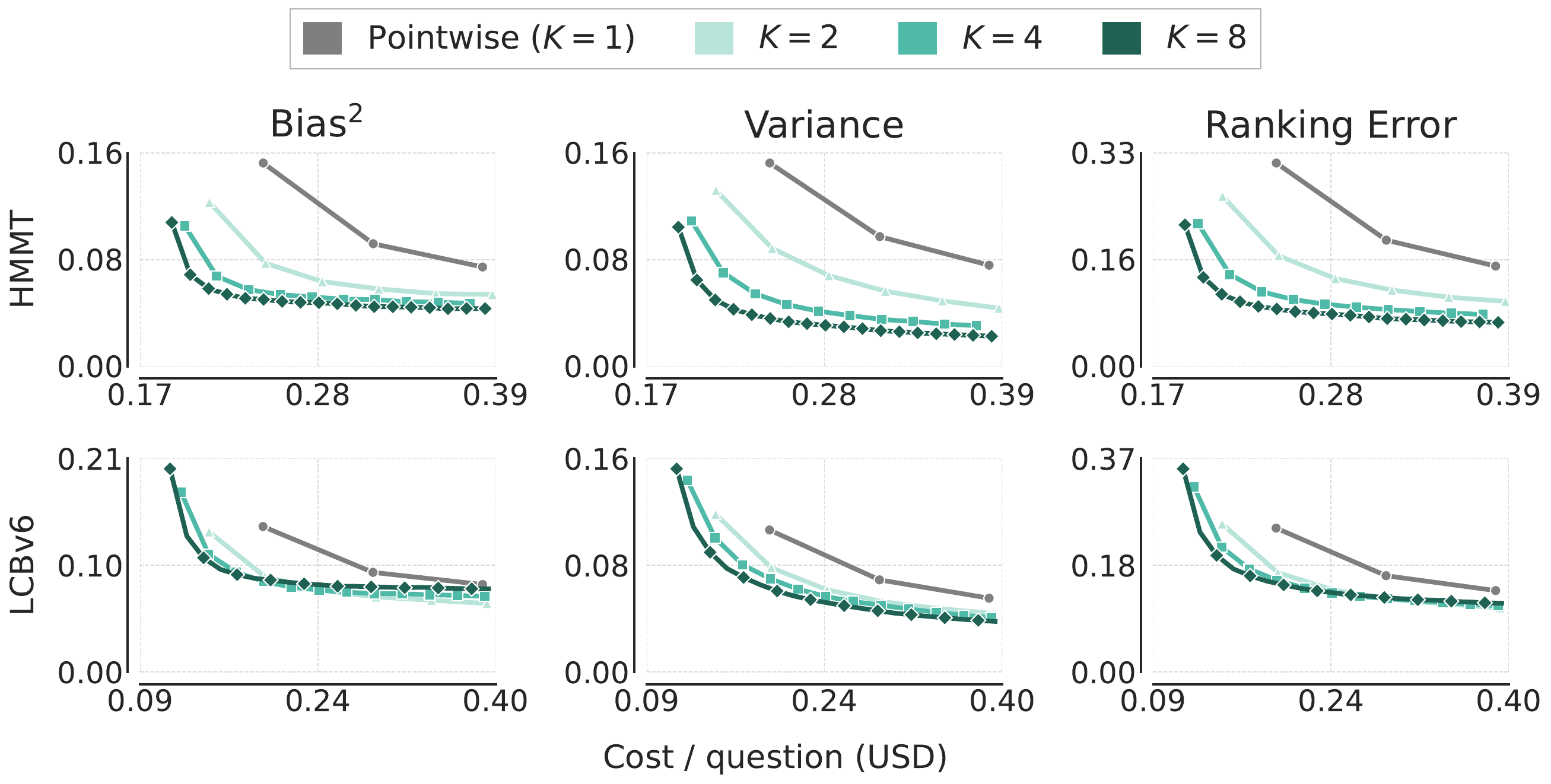}
  \caption{\textbf{Larger groups reduce variance faster but can increase bias.} Results for GPT-OSS-20B on HMMT (math) and LCBv6 (code). Increasing the group size allows more evaluations within the same cost budget, reducing variance faster. However, these gains can come at the expense of increased bias: on LCBv6, $K=8$ yields higher bias than smaller groups. Similar trends are observed for Qwen3.5-4B (see \autoref{fig:kladder_qwen}).}
  \label{fig:kladder}
\end{figure}

\textbf{Joint evaluation reduces ranking error, but larger groups can increase bias.}
Consistent with \citet{singh2026v_1} and \citet{kwok2026llmasaverifiergeneralpurposeverificationframework}, we find that evaluating candidates together improves ranking quality over evaluating them in isolation. In \autoref{fig:kladder}, we vary the group size $K\in\{1,2,4,8\}$ on HMMT and LCBv6 while keeping the grouping strategy random. Pointwise scoring ($K=1$) has higher bias than joint evaluation, suggesting that comparisons help the model distinguish correct solutions from incorrect ones (see \autoref{app:groupwise-comp-bias} for an example). Having multiple candidates in the context also reduces variance because fewer calls are needed per scoring round, allowing each candidate's score to be averaged over more evaluations within the same cost budget. However, these variance reductions can come at the expense of bias: on LCBv6, increasing the group size to $K=8$ raises bias, potentially because comparing more candidates in a single context is harder. We therefore use $K=4$ as the default to balance these effects.

\begin{figure}[t]
  \centering
  \includegraphics[width=\textwidth]{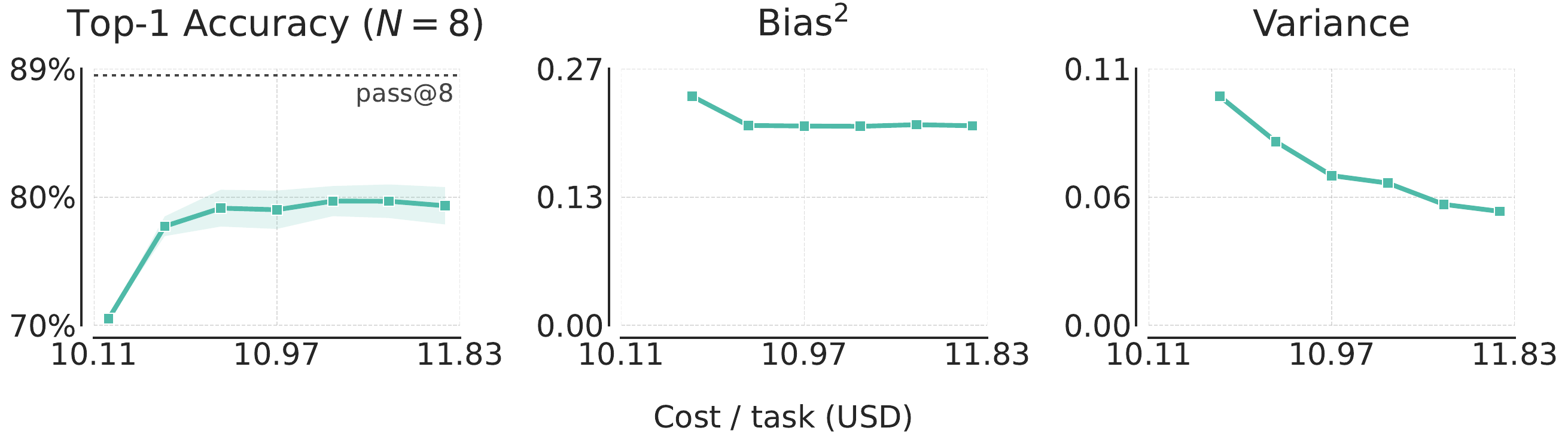}
  \caption{\textbf{Residual bias persists at scale with Gemini-3.1-Pro on Terminal-Bench-2.1.} We generate $N=8$ candidates and score them with \scorpio. Top-1 accuracy for \scorpio ($K=4$) plateaus about $9$ percentage points below $\mathrm{pass@}8$. \scorpio's bias plateaus with compute, while its variance continues to decrease. Results are averaged across $8$ seeds.}
  \label{fig:tb_3panel}
\end{figure}

\textbf{Residual bias persists at frontier scale.}
To examine whether the limitations of self-verification persist in a more capable model, we evaluate Gemini-3.1-Pro~\citep{google2026gemini31pro} on Terminal-Bench-2.1~\citep{merrill2026terminal}. We generate $N=8$ responses per task and use \scorpio to select the best candidate by comparing candidate summaries. As shown in \autoref{fig:tb_3panel}, \scorpio's selection accuracy plateaus $9$ percentage points below $\mathrm{pass@}8$. The middle and right panels show a similar trend: increasing inference compute reduces variance, while a residual bias remains. These results show that simple scoring methods can improve selection accuracy even at frontier scale and suggest that reducing residual ranking bias could help close the remaining gap.

\subsection{Test-Time Scaling for Tree Search}
\label{sec:tree_search}

Now we apply \scorpio to tree search to score partial solutions and identify promising intermediate trajectories. We use DeepConf’s search procedure~\citep{fu2025deepthinkconfidence}, replacing its confidence-based evaluation of partial trajectories with \scorpio's scoring. The model evaluates groups of partial solutions and scores how likely each is to lead to a correct final answer (see~\autoref{alg:tree-search}). Intermediate solutions are periodically evaluated every $2{,}048$ tokens. After each evaluation, the lowest-scoring partial solutions are pruned before they complete, so no further generation is spent on unpromising trajectories. We compare \scorpio with ThinkPRM~\citep{khalifa2025thinkprm}, which uses a trained process reward model to score partial solutions pointwise, and with DeepConf.

In this setup, compute can be scaled along two axes: by increasing the number of initial solutions and by increasing the number of times each partial soltuion is evaluated before pruning. For \scorpio, we vary the initial pool size $N\in\{8,32,128\}$, use groups of size $K=4$, and repeat the random grouping and scoring process three times ($M=3$). At each pruning stage, the top $\rho=0.75$ fraction of partial solutions is retained, while the rest are pruned. 
For each baseline, we follow the final-answer selection strategy reported in its original paper, with any adaptations described in \autoref{app:tree-search}.

\begin{figure}[t]
  \centering
    \includegraphics[width=0.85\textwidth]{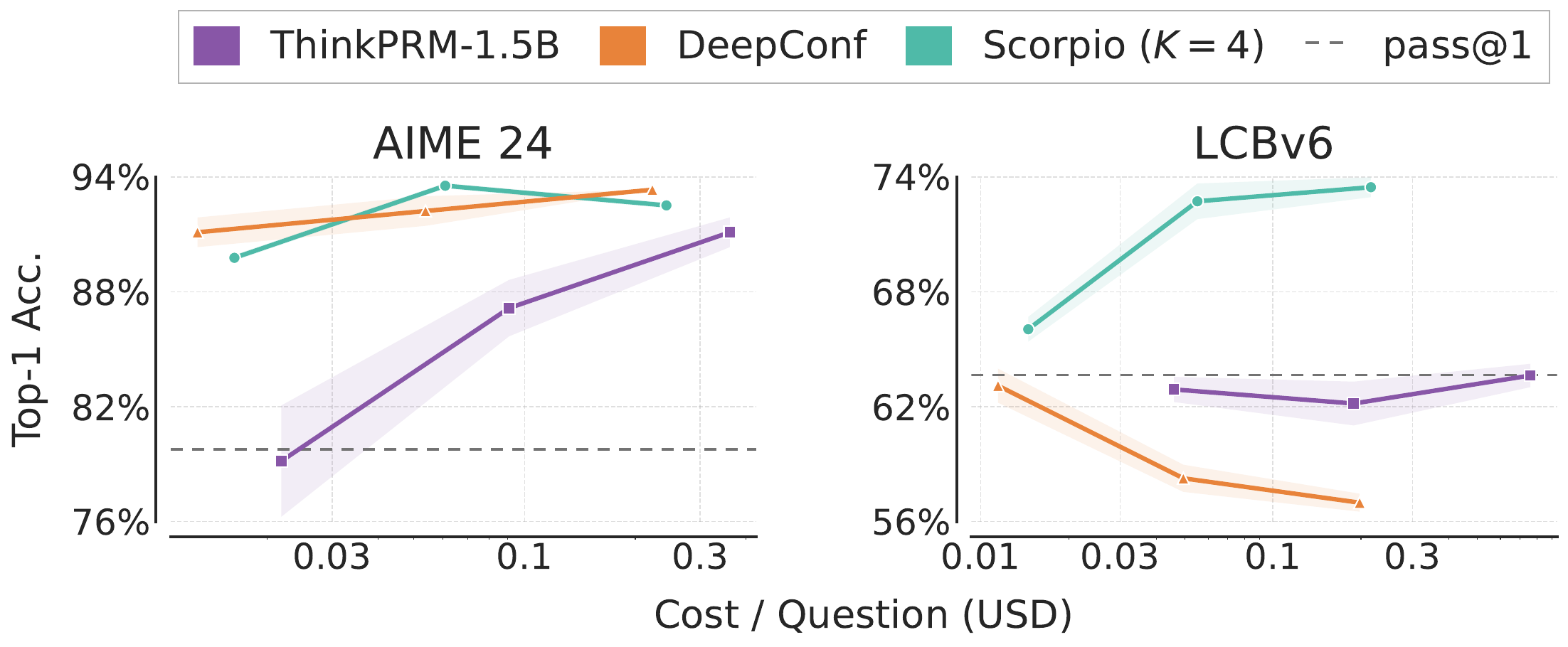}
  \caption{\textbf{Test-time search using \scorpio matches or outperforms prior methods without any training.} Results for GPT-OSS-20B on AIME 2024 and LCBv6. We compare \scorpio, which jointly scores partial solutions during search, with ThinkPRM-1.5B ~\citep{khalifa2025thinkprm}, which uses a trained process reward model, and DeepConf, which uses confidence scores derived from token probabilities. We vary the number of initial solutions: $N\in \{8,32,128\}$ for \scorpio and ThinkPRM, and $N\in \{16,64,256\}$ for DeepConf to keep total costs comparable. Top-1 accuracy is plotted against total inference cost. Similar trends are observed for Qwen3.5-4B (see \autoref{fig:value_nscaling_qwen}).}
  \label{fig:value_nscaling}
\end{figure}

\textbf{\scorpio improves test-time search on math and code.}
\autoref{fig:value_nscaling} compares \scorpio, DeepConf, and ThinkPRM on AIME 2024 and LCBv6 using GPT-OSS-20B. We vary $N\in\{8,32,128\}$ for \scorpio and ThinkPRM, while allowing DeepConf larger pools of $N\in\{16,64,256\}$ to keep total costs in comparable ranges. On AIME 2024, where the model already achieves a high $\mathrm{pass@}1$, DeepConf and \scorpio attain similar peak accuracy, suggesting that inexpensive confidence estimates can be effective when models are already good at solving the task. On LCBv6, where $\mathrm{pass@}1$ is lower, \scorpio scales well with compute, exceeding the strongest baseline's peak accuracy by \textbf{10} percentage points. We believe that ThinkPRM's lower accuracy on code is primarily due to a distribution mismatch between its mathematical training data and the coding tasks evaluated here.

\textbf{Repeated evaluation of partial solutions improves search accuracy.} We now fix the number of starting solutions at $N=128$ and vary the number of times each partial solution is scored, $M\in\{1,2,3\}$. As shown in \autoref{fig:m_scaling}, across three datasets and two models, spending more compute on state evaluation improves the accuracy of the selected answer by \textbf{2--6} percentage points.

These results show that search with \scorpio can match or outperform ThinkPRM and DeepConf, and that additional compute spent evaluating partial solutions can improve search accuracy. However, as in verification, a gap to $\mathrm{pass@}N$ persists
after search.

\subsection{Test-Time Scaling for Aggregation}
\label{sec:aggregation}
Finally, we evaluate test-time scaling via aggregation, which iteratively generates $N$ candidates and uses them as prompt-conditioning to generate $N$ new, improved candidates. Following \citet{venkatraman2025rsa}, we scale across $L=1 \to 10$ rounds of generation and aggregation, and further scale compute by varying the number of candidates generated at each round, $N\in\{8,32,128\}$.

We compare a standard aggregation method RSA~\citep{venkatraman2025rsa} with RSA augmented by \scorpio scoring. Following \citet{singh2026v_1}, \scorpio scores the $N$ candidates after each round of generation, and only the top $N/2$ are kept for prompt-conditioning: the $N$ inputs to the LLM in the next round are built from subsets randomly sampled from these top $N/2$ candidates.

\begin{figure}[t]
  \centering
  \includegraphics[width=\textwidth]{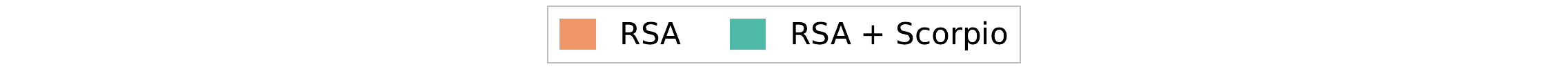}\\[-0.3em]
  \begin{subfigure}[t]{0.577\textwidth}
    \includegraphics[width=\textwidth]{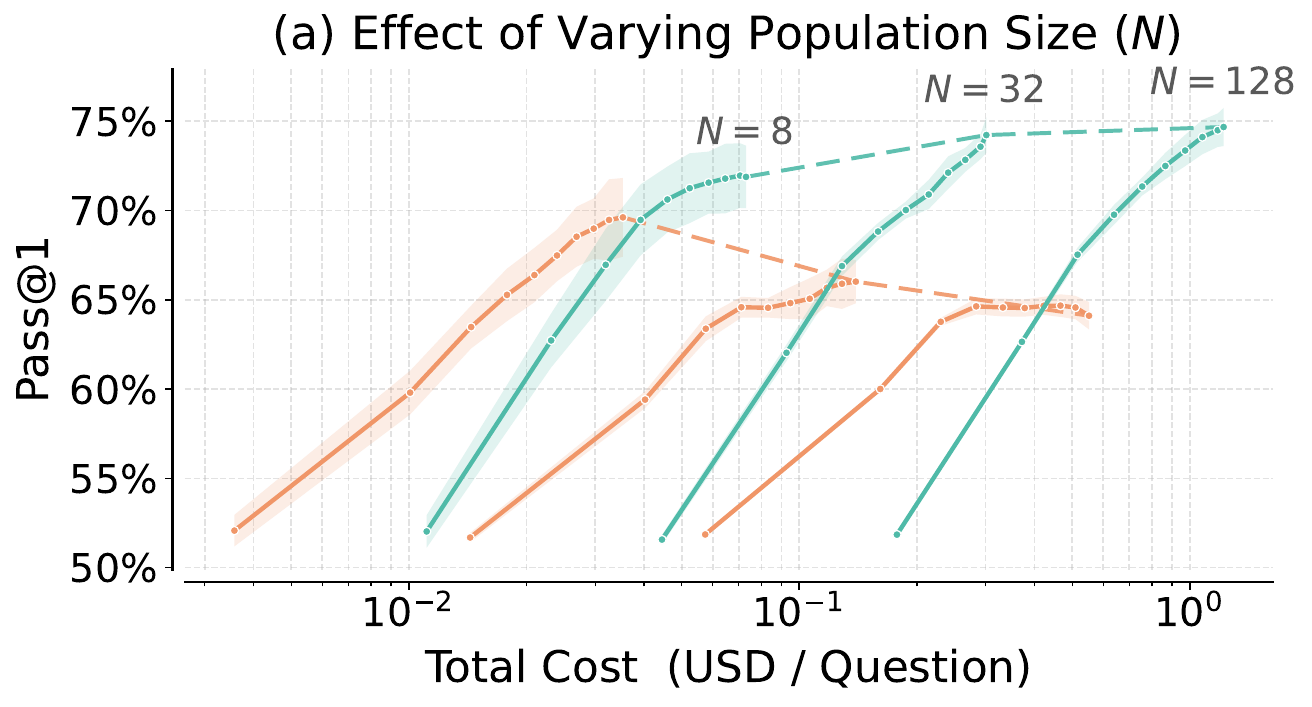}
    \phantomsubcaption\label{fig:rsa_scorpio_a}
  \end{subfigure}\hfill
  \begin{subfigure}[t]{0.403\textwidth}
    \includegraphics[width=\textwidth]{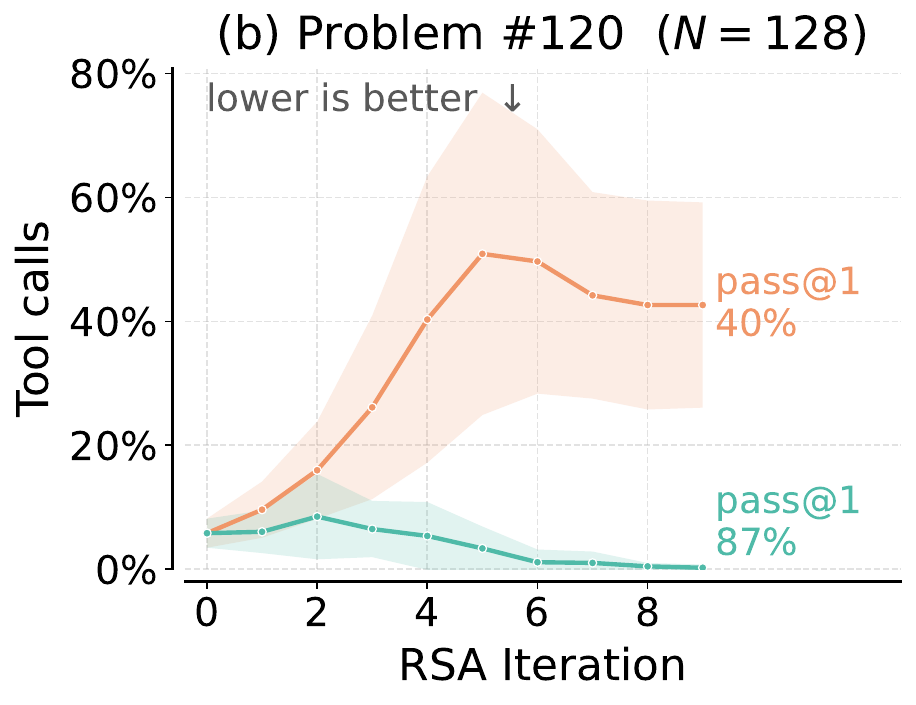}
    \phantomsubcaption\label{fig:rsa_scorpio_b}
  \end{subfigure}
  \caption{\textbf{Pruning candidates with \scorpio allows RSA to scale with pool size $N$.} Using GPT-OSS-20B on LCBv6, we plot \textbf{(a)} accuracy across $L=1\to10$ aggregation rounds for different pool sizes $N$. Increasing $N$ degrades RSA performance, whereas RSA + \scorpio improves with larger pools. One failure mode contributing to this degradation is shown in \textbf{(b)}: RSA amplifies rare incorrect responses over successive rounds on problem 120. LCBv6 requires a complete Python solution in a single turn and does not permit tool calls. However, RSA amplifies rare incorrect responses containing tool calls over successive rounds, with the effect becoming more severe as $N$ increases (see \autoref{fig:app_rsa_toolcalls}). After each round, \scorpio scores all $N$ candidates, and only the top $N/2$ are used to construct aggregation sets for the next round, limiting this bad-mode amplification. Results are averaged over 8 seeds, with $95\%$ CI.}
  \label{fig:rsa_scorpio_scaling}
\end{figure}

\textbf{Standard RSA can degrade as the number of candidates $N$ increases.} In \autoref{fig:rsa_scorpio_a}, we scale RSA over $L=1 \to 10$ rounds and candidate pool sizes $N \in \{8,32,128\}$. Performance improves over rounds $L$, but scaling compute by increasing $N$ from $8$ to $128$ lowers final accuracy by \textbf{6} percentage points despite using around $16\times$ more compute.
To illustrate how this decline can arise, we examine a single LCBv6 problem (\#120). Although the model is instructed to provide a complete Python solution in a single turn, it occasionally attempts tool calls to test its logic instead. These responses fail because the evaluation harness expects a complete program and does not support tool calls. When these tool-call responses are passed into subsequent aggregation rounds, this behavior becomes more common, despite instructions to provide a Python program. With $N=128$, the proportion of responses containing tool calls grows from $5.8\%$ of the initial pool to $40\%$ of the final pool, shown in \autoref{fig:rsa_scorpio_b}. For this problem, tool-call responses remain rare with $N=8$ but become substantially more common with $N=128$ (see \autoref{fig:app_rsa_toolcalls} in Appendix).

\textbf{RSA + \textsc{Scorpio} improves scaling with $N$ and limits the amplification of undesirable responses.} RSA + \scorpio selects the top $N/2$ candidates at each round to maintain a high-quality pool for the next round. For the problem examined in \autoref{fig:rsa_scorpio_b}, this selection reduces tool-call responses to nearly zero and improves final accuracy from $40\%$ to $87\%$. \scorpio's scoring procedure incurs extra compute cost but allows RSA to scale well with $N$. As shown in \autoref{fig:rsa_scorpio_a}, RSA + \scorpio achieves a better compute--accuracy tradeoff and higher peak accuracy within the evaluated budgets.

\section{Conclusion}
\label{sec:conclusion}

In this work, we introduced a bias-variance decomposition to analyze how well LLMs evaluate their own solutions. We found that increasing test-time compute reduces variance, while residual ranking bias limits further improvements in selection accuracy. Using this decomposition, we identified strengths and weaknesses of test-time scaling algorithms and introduced \scorpio, a simple strategy that jointly scores candidates over random partitions. This strategy is effective across multiple test-time scaling settings, models, and benchmarks. More broadly, the decomposition provides a novel perspective on compute and algorithmic design choices and encourages future test-time scaling methods to improve judgments by reducing bias, variance, or both. A preliminary experiment in \autoref{app:cross-model} points to one such direction: replacing the self-judgments of Qwen3.5-4B with those of other models can reduce ranking bias. Using judges whose biases differ from those of the model generating the solutions may therefore be a promising way to reduce residual ranking bias.

\subsection*{AI use statement}

In this work, we used generative AI tools to polish the writing, improve the style of plots, and help with the implementation details of our method. We did not use these tools to generate ideas or propose experiments. We reviewed all AI-assisted work and verified that grammatical changes preserved the intended meaning. We take responsibility for the final content of this work, including all text, claims, and artifacts produced with the aid of generative AI.

\subsection*{Reproducibility statement}

\autoref{app:exp-setup} provides details of our method and the baselines used throughout the paper. \autoref{app:prompts} provides details of the prompts used for self-evaluation. Our code, together with instructions to reproduce our experiments, is available at \url{https://github.com/anirudhb11/scorpio}.

\subsection*{Author contributions}
AB proposed measuring ranking error as a function of inference compute and scoring candidates with groupwise comparisons, implemented all experiments in the paper, and wrote the initial draft.

KR ran the experiments on scaling test-time compute for verification with GPT-OSS-20B, the experiments on aggregation, the cross-model judge experiments, and some of the Qwen3.5-4B experiments on LCBv6. SY ran the experiments on scaling test-time compute with Qwen3.5-4B on HMMT Feb 2025 and Reasoning Gym. MMM ran the DeepConf experiments on LCBv6 and took part in project discussions. JOC ran the cross-model judge experiments with the Qwen3.5-9B judge, provided feedback on experiments, and helped with refining the draft. KR, SL, and SY also ran an initial version of the tree-search experiments.

SL, MN, LC, and AC advised the project. Suggestions from SL and MN shaped the direction of the project: multiple discussions with SL scoped the project down to test-time scaling, and MN suggested experiments to show that the approach generalizes across test-time scaling methods, from verification to aggregation and search. SL and MN also improved the presentation of results and rewrote parts of the draft. LC and AC helped frame the research questions, gave feedback on experiments, and edited the draft.

\subsection*{Acknowledgements}
AB thanks Jatin Prakash, Ankur Sikarwar, Ayush Agarwal, Madhur Panwar, and Shubham Tiwari for helpful discussions and suggestions. This work was supported by the Institute of Information \& Communications Technology Planning \& Evaluation (IITP) grant funded by the Korean government (MSIT) (No. RS-2024-00457882, National AI Research Lab Project), the CIFAR AI Chair Program, and a gift from Qualcomm. We thank the Mila IDT team for their technical support. This research was enabled in part by compute resources provided by Mila (mila.quebec), Calcul Qu\'ebec (calculquebec.ca), and the Digital Research Alliance of Canada (alliancecan.ca). Finally, we thank Google for awarding Gemini API credits for this project through the Gemini Academic Program.

\bibliographystyle{main}
\bibliography{references}

\appendix
% Appendix only: let text share pages with figures, and stack figure-only pages from the top.
\renewcommand{\topfraction}{0.9}
\renewcommand{\bottomfraction}{0.8}
\renewcommand{\textfraction}{0.07}
\renewcommand{\floatpagefraction}{0.8}
\setcounter{topnumber}{3}
\setcounter{totalnumber}{4}
\makeatletter
\setlength{\@fptop}{0pt}
\setlength{\@fpsep}{12pt plus 2fil}
\setlength{\@fpbot}{0pt plus 1fil}
\makeatother
\section{Limitations}
\label{sec:limitations}

Despite \scorpio's improved test-time scaling, it does not eliminate bias in rankings, as a result of which a significant gap remains between its selection accuracy and the oracle accuracy, $\mathrm{pass@}N$. Addressing this residual bias remains an important direction for realizing the full benefit of sampling multiple candidates.

\section{Extended Related Work}
\label{app:ext-rel-work}
\textbf{Repeated Groupwise Comparisons using LLM Judges. } 
Prior work has studied joint scoring and repeated evaluation for text quality assessment \citep{yuan-etal-2024-batcheval} and passage relevance ranking \citep{korikov-etal-2025-batched}, finding improvements in judgment quality and efficiency. \scorpio shares this principle of joint scoring with randomized contexts. However, our work differs in its objective, analysis, and applications. First, we study whether a model can identify correct solutions among its own generations, rather than rank passages by relevance to a query. Second, we use a bias and variance decomposition of ranking error to analyze repeated evaluation and grouping strategies. This analysis sheds light on how inference compute affects both the consistency and correctness of self-evaluation, and how these effects depend on algorithmic design choices. Third, we use \scorpio to guide subsequent generation, beyond ranking a fixed candidate pool: during search, it selects partial solutions to continue; during aggregation, it selects solutions to condition the next round of generation.

\textbf{Privileged Information for Evaluating Candidates. }
Prior work on training LLMs using actor-critic methods has shown that critics produce better value estimates when given richer context: \citet{venkatraman2026critique} condition the critic on rollouts and rewards of other trajectories in a group, and \citet{wen2026trust} condition on a correct reference solution. We find an analogous benefit at inference time, where conditioning on solutions of other candidates improves the ability to identify correct solutions. Complementary to these works, we examine how design choices, such as the number of candidates evaluated in a single LLM call and how they are grouped, affect evaluation quality.

\section{Ranking Error and Bias Variance Decomposition}

\subsection{Estimation of Bias and Variance for Different Methods}
\label{app:bias-var-computation}

We estimate the bias and variance from $R=20$ rankings at a given evaluation budget. For each incorrect--correct
pair $(i,j)$, we first estimate the probability of an incorrect
ordering, then average the resulting squared bias and variance
over all such pairs:

\begin{align}
\hat{p}_{ij}
&=
\frac{1}{R}\sum_{r=1}^{R}
\mathbbm{1}\!\left[
\hat{s}^{(i,r)} \geq \hat{s}^{(j,r)}
\right],
\qquad
y^{(i)}\in\gC^-,\quad y^{(j)}\in\gC^+.
\label{eq:pairwise-error-estimate}
\\
\widehat{\mathrm{Bias}^2}
&=
\frac{1}{|\gC^+||\gC^-|}
\sum_{y^{(i)}\in\gC^-}
\sum_{y^{(j)}\in\gC^+}
\left[\hat{p}_{ij}^{\,2}-\frac{\hat{p}_{ij}(1-\hat{p}_{ij})}{R-1}\right],
\label{eq:bias-estimate}
\\
\widehat{\mathrm{Variance}}
&=
\frac{1}{|\gC^+||\gC^-|}
\sum_{y^{(i)}\in\gC^-}
\sum_{y^{(j)}\in\gC^+}
\frac{R}{R-1}\,\hat{p}_{ij}(1-\hat{p}_{ij}).
\label{eq:variance-estimate}
\end{align}

where $\hat{s}^{(i,r)}$ is the score of candidate $y^{(i)}$
in the $r$-th ranking at the evaluation budget under consideration. Because $\hat{p}_{ij}$ averages $R$ Bernoulli outcomes, $\E[\hat{p}_{ij}^{2}]=p_{ij}^{2}+p_{ij}(1-p_{ij})/R$ and $\E[\hat{p}_{ij}(1-\hat{p}_{ij})]=\frac{R-1}{R}\,p_{ij}(1-p_{ij})$, so the plug-in estimates $\hat{p}_{ij}^{2}$ and $\hat{p}_{ij}(1-\hat{p}_{ij})$ would overestimate the squared bias and underestimate the variance. The corrections in \autoref{eq:bias-estimate} and \autoref{eq:variance-estimate} make both estimators unbiased, and the two still sum to $\hat{p}_{ij}$. Algorithm~\ref{alg:sequential-bias-var} describes how these rankings are obtained.

\begin{algorithm}[t]
\caption{Estimating bias and variance}
\label{alg:sequential-bias-var}
\begin{algorithmic}[1]
\Require Problem $x$, labeled candidate pool $\gC$ of size $N$,
scorer $f$, maximum evaluation level $L_{\max}$,
target level $1\leq L\leq L_{\max}$
\State Set $R=20$ independent runs
\State Initialize score array
$S\in\mathbb{R}^{R\times L_{\max}\times N}$

\For{$r=1,\ldots,R$}
    \State Initialize a fresh run of $f$ on $(x,\gC)$
    \For{$\ell=1,\ldots,L_{\max}$}
        \State Advance $f$ to evaluation level $\ell$
        \Comment{Continue the same run using its scoring history}
        \State Store the current candidate scores in $S[r,\ell,:]$
    \EndFor
\EndFor

\For{$r=1,\ldots,R$}
    \State $(\hat{s}^{(1,r)},\ldots,\hat{s}^{(N,r)})
    \gets S[r,L,:]$
    \Comment{Retrieve scores at the target level}
\EndFor
\State Compute $\hat{p}_{ij}$ using
~\eqref{eq:pairwise-error-estimate}
\State Compute squared bias and variance using
~\eqref{eq:bias-estimate} and~\eqref{eq:variance-estimate}
\State \Return Estimated squared bias and variance
\end{algorithmic}
\end{algorithm}

\subsection{Groupwise Comparisons Lower Bias Relative to Pointwise Scoring}
\label{app:groupwise-comp-bias}

Here, we examine responses to a single question to illustrate the benefit of \scorpio's groupwise comparisons over pointwise evaluations. All excerpts below are from GPT-OSS-20B generations for problem index $0$ in the HMMT-Feb 2025 dataset. In the verification setup, we start with $N=128$ generations for this question. Of these, $45$ yield the correct answer, while $67$ yield the same incorrect answer. Thus, majority voting also does not select the correct answer.

\begin{greybox}{Question and Answer}
\small
Given that $x$, $y$ and $z$ are positive real numbers such that
\[
x^{\log_2(yz)}=2^{8}\cdot 3^{4},\qquad
y^{\log_2(zx)}=2^{9}\cdot 3^{6},\qquad
z^{\log_2(xy)}=2^{5}\cdot 3^{10},
\]
compute the smallest possible value of $xyz$.
\tcblower
\small
\textbf{Answer:} $\frac{1}{576}$.
\end{greybox}

\textbf{Failure mode.} Two solutions satisfy the constraints, one yielding $\frac{1}{576}$ (correct as it is smaller) and the other $576$ (incorrect). They are related by reversing the signs of $\log_2 x$, $\log_2 y$, and $\log_2 z$. The incorrect reasoning begins by assuming $x,y,z>1$, even though the question only requires $x,y,z$ to be \textit{positive}. This assumption excludes the negative logarithmic branch and overlooks the solution that yields $\frac{1}{576}$.

\textbf{Pointwise evaluation does not spot the mistake.} We track one such incorrect generation and score it $20$ times under each scheme. Scored on its own it receives a rating of $10$ on $19$ of $20$ evaluations (mean $9.95$); scored inside random groups of $K=4$ it receives a $1$ or a $2$ on $17$ of $20$ (mean $2.60$).

\textbf{Groupwise evaluation spots the mistake.} We now show one \scorpio call in which
this generation is scored alongside three others. Solutions 1, 2 and 3 all answer $576$ (incorrect one),
while Solution 4 answers $1/576$ (correct). The verifier
opens by comparing the final answers, and identifies the faulty assumption directly:

\begin{greybox}{Verifier reasoning}
\small
``We need to see whether the smallest possible product is $576$ or $1/576$. \ldots{} But
$x,y,z$ must be positive real numbers. Negative exponents produce positive numbers: e.g.,
if $X=-2$, then $x=2^{-2}=1/4$ positive. So that's okay. \ldots{} but the requirement says
$x,y,z$ positive real numbers, \emph{not necessarily $>1$}. So negative $X$ allowed.''
\end{greybox}

\noindent
It then verifies that the negative branch satisfies the original equations, and assigns a
rating of $1$ to Solutions 1--3 and $10$ to Solution~4. In other words, performing groupwise comparisons helps reduce the bias while evaluating solutions.

\subsection{Why Does Bias Initially Decrease with Compute?}
\label{app:bias-dec}

\begin{figure}[htbp]
  \centering
  \includegraphics[width=0.78\textwidth]{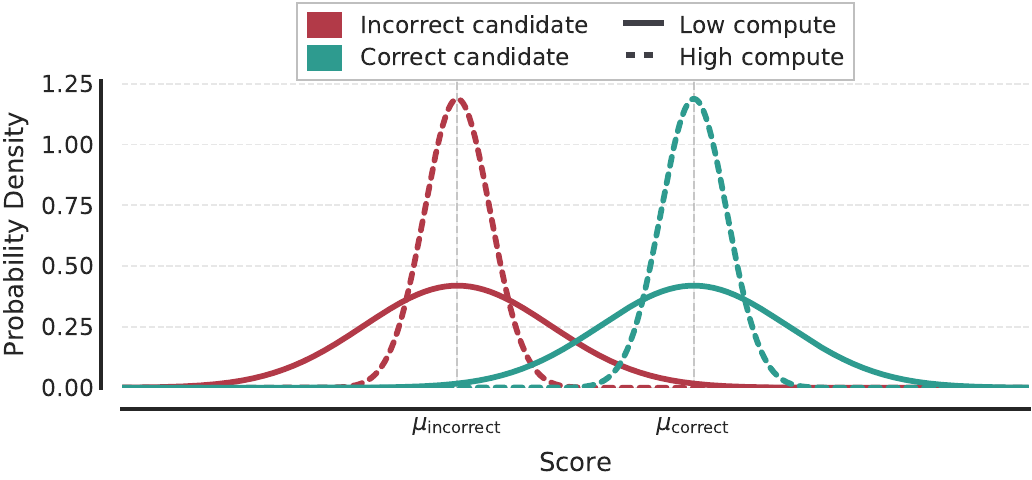}
  \caption{\textbf{Increasing compute does not change the mean of the score distribution, but can reduce the bias in rankings.} At low compute (solid curves), overlapping score distributions allow the incorrect candidate to receive a higher score than the correct one. Averaging more evaluations (dotted curves) concentrates the scores around their respective means. Since the correct candidate has a higher mean score in this example, this makes an incorrect ordering less likely, reducing the pair's squared bias $(\E[Z_{ij}])^2$.}
  \label{fig:score-concentration}
\end{figure}

As seen in Figure~\ref{fig:main_biasvar}, bias initially
decreases as inference compute increases. To understand
why, consider a correct and an incorrect candidate whose
score distributions overlap, with the correct candidate
receiving a higher expected score, as illustrated in
\autoref{fig:score-concentration}. At a low evaluation
budget, fluctuations in these scores can cause the
incorrect candidate to be ranked above the correct one,
despite its lower expected score.

As the evaluation budget increases, each candidate's final
score averages more judgments and concentrates around its
mean. This makes an incorrect ordering less likely,
reducing the pair's contribution $(\E[Z_{ij}])^2$ to the
bias term even though the expected scores remain unchanged.
Averaging does not, however, correct pairs for which the
incorrect candidate has a higher expected score. For these
pairs, concentrating the scores makes the incorrect
ordering more likely, contributing to the bias floor
observed at higher budgets.

\subsection{Residual Bias Dominates Ranking Error at High Compute}
\label{app:var-rate}

\begin{figure}[htbp]
  \centering
  \includegraphics[width=\textwidth]{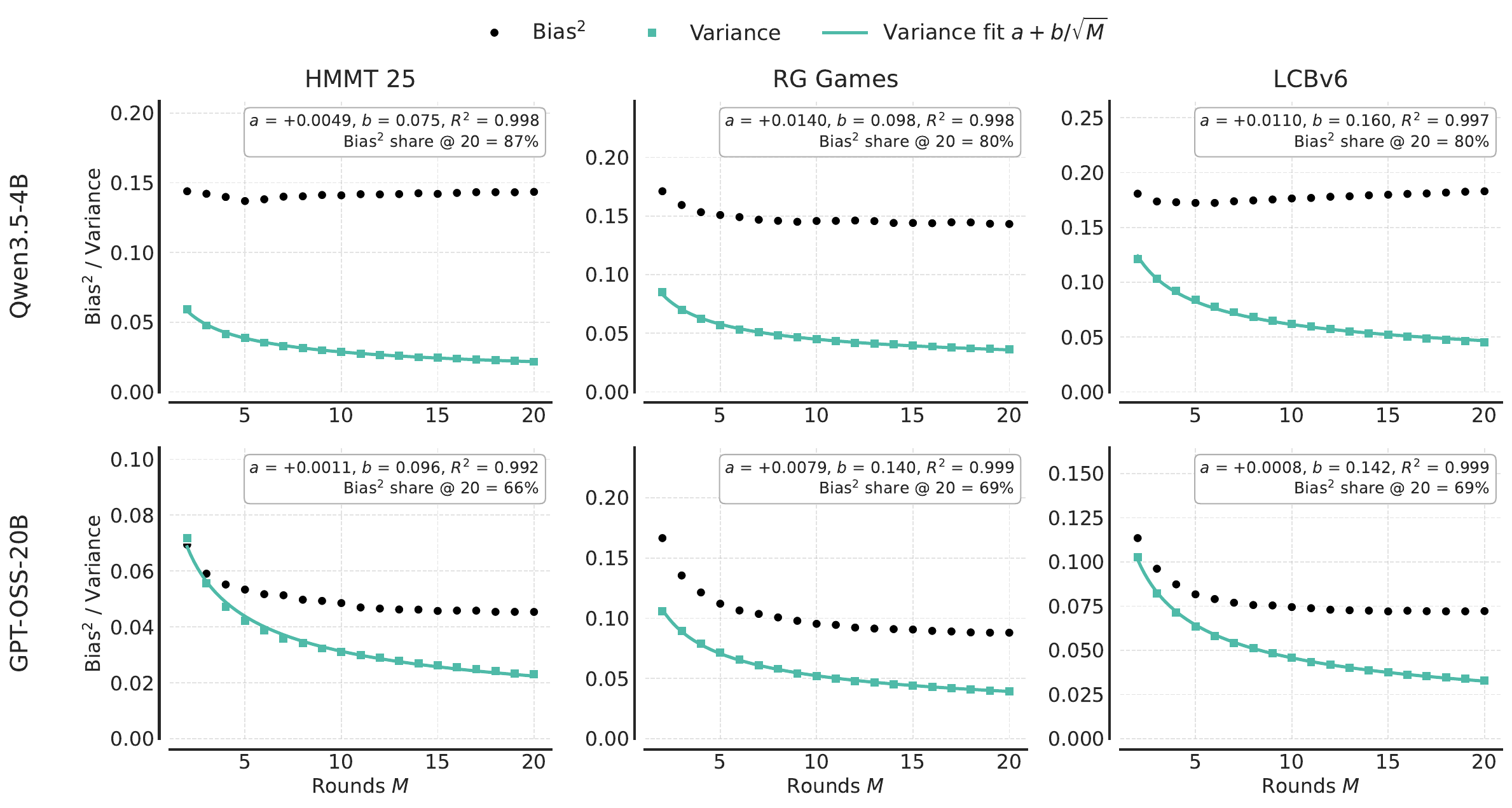}
  \caption{\textbf{Residual bias dominates ranking error as compute grows.} Variance and squared bias of the ranking error of \scorpio ($K=4$) versus the number of scoring rounds $M$ for Qwen3.5-4B (top) and GPT-OSS-20B (bottom). Black circles show the squared bias, which flattens as $M$ grows, and teal squares the variance; curves show least-squares fits of $a+b/\sqrt{M}$ to the variance. The box gives the fitted $a$, $b$, and $R^2$ for the variance, and the share of the ranking error due to squared bias at $M=20$. Using the asymptotic variance $a$ together with the squared bias at $M=20$, we estimate that bias would account for $91\%$ to $99\%$ of the ranking error as $M$ grows.}
  \label{fig:variance_decay}
\end{figure}

\autoref{fig:variance_decay} shows how the variance and squared bias of \scorpio ($K=4$) change with the number of scoring rounds $M$ for both models on all three datasets. Fitting $\mathrm{Var}(M)=a+b/\sqrt{M}$ by least squares to the estimated variances yields a strong fit, with $R^2$ between $0.992$ and $0.999$, so variance decreases approximately as $1/\sqrt{M}$. Squared bias, in contrast, flattens: between $M=10$ and $M=20$ it changes by at most $8\%$, while variance drops by $19\%$ to $28\%$. At $M=20$, bias already accounts for $66\%$ to $87\%$ of the ranking error. The fitted intercept $a$ estimates the asymptotic variance as $M\to\infty$. Using $a$ together with the squared bias at $M=20$, we estimate that bias would account for $91\%$ to $99\%$ of the ranking error as $M$ grows, highlighting residual bias as the primary contributor to ranking error.

\subsection{Cross-Model Judgements}
\label{app:cross-model}

The results in the main text highlight two findings: (i) the design of self-verification pipelines can be simplified, as random grouping matches or outperforms more sophisticated tournament-based approaches, and (ii) despite these improvements, a considerable gap remains between the accuracy of the selected answer and $\mathrm{pass@}N$. As shown in \autoref{app:var-rate}, a key contributor to this gap is ranking bias, which stops improving as compute increases.

\begin{figure}[htbp]
  \centering
  \includegraphics[width=\textwidth]{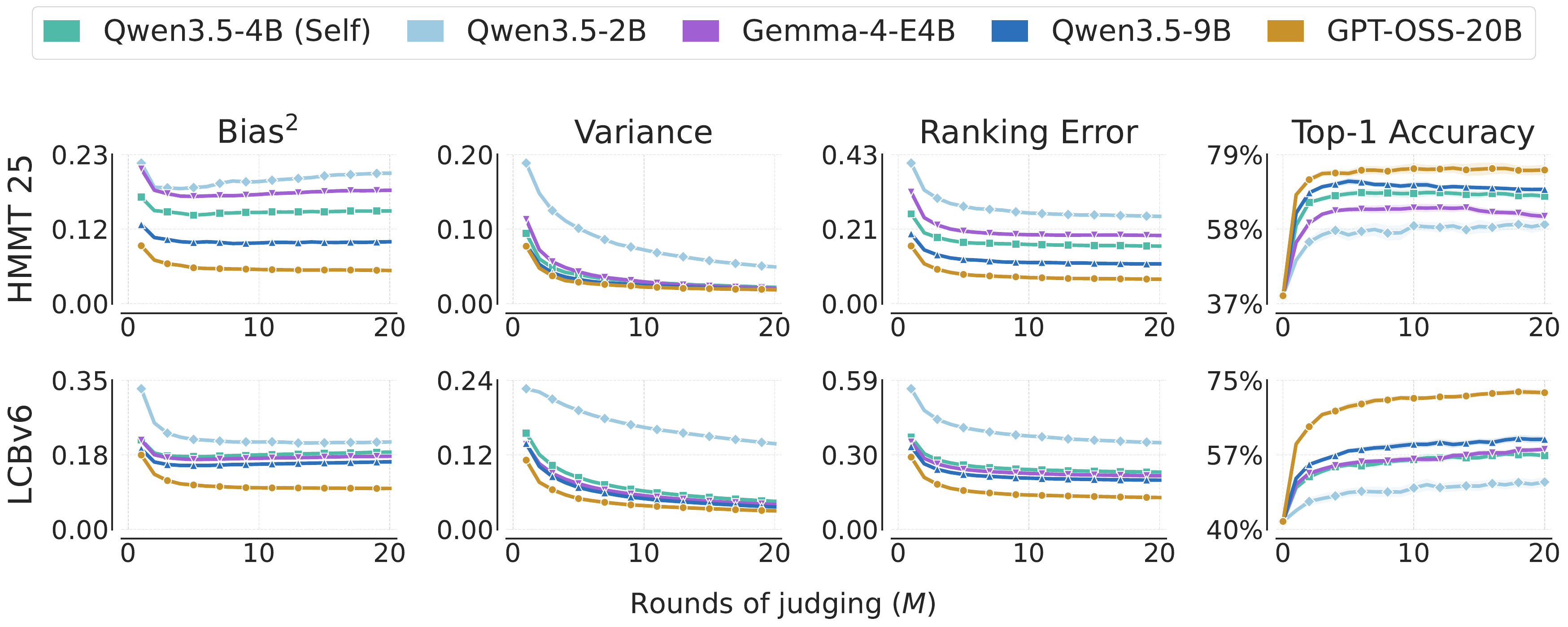}
  \caption{\textbf{A judge from a different model can reduce ranking bias.} Candidates generated by Qwen3.5-4B are scored with \scorpio ($K=4$) either by \textbf{Qwen3.5-4B itself (the self judge)} or by one of four other judges, on HMMT Feb 2025 (top) and LCBv6 (bottom). We plot bias$^2$, variance, ranking error, and top-1 accuracy as the number of scoring rounds $M$ increases, averaged over $20$ repetitions; shaded regions show $95\%$ confidence intervals for top-1 accuracy. At $M=20$, replacing the self judge with GPT-OSS-20B reduces the ranking error from $0.17$ to $0.07$ on HMMT and from $0.23$ to $0.13$ on LCBv6, and improves top-1 accuracy from $67.6\%$ to $75.0\%$ and from $57.0\%$ to $71.6\%$, respectively.}
  \label{fig:crossjudge}
\end{figure}

To mitigate this bias, we consider a setting in which one model generates the solutions and a different model evaluates them. The use of multiple models for efficient test-time scaling has been explored in \citet{maheswaran2026squeeze}, where the primary focus has been to improve the trade-off between cost and accuracy by assigning models of different strengths to different stages. Here, we examine the use of multiple models to reduce ranking bias. We fix the generating model to Qwen3.5-4B and consider four judges: (i) Qwen3.5-2B, (ii) Gemma-4-E4B, (iii) Qwen3.5-9B, and (iv) GPT-OSS-20B. \autoref{fig:crossjudge} reports the squared bias, variance, ranking error, and top-1 accuracy of each judge as the number of scoring rounds $M$ increases. More capable judges, from the same model family (Qwen3.5-9B) or a different one (GPT-OSS-20B), reduce bias and improve top-1 accuracy over the self judge on both datasets.

\begin{figure}[htbp]
  \centering
  \includegraphics[width=\textwidth]{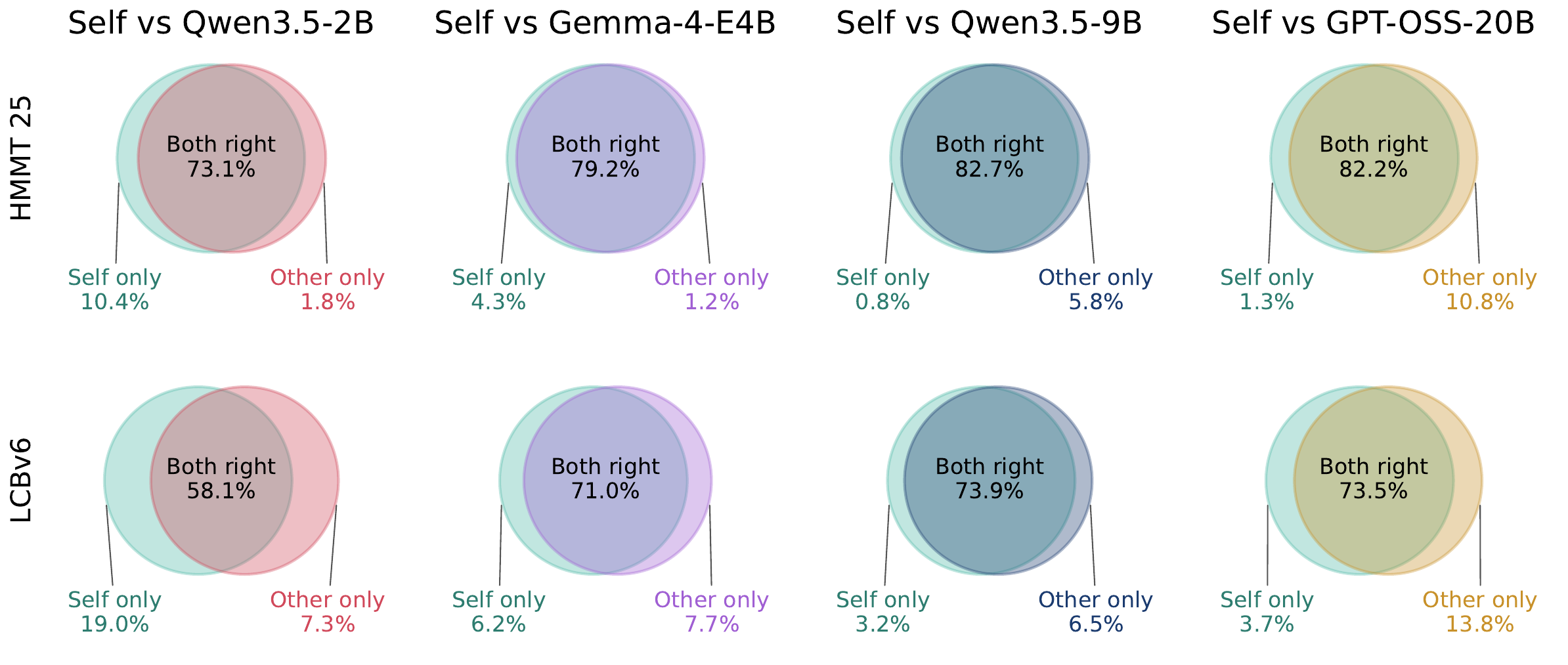}
  \caption{\textbf{Different models have different biases.} For every pair of a correct and an incorrect candidate generated by Qwen3.5-4B, we record whether the self judge and another judge rank the correct candidate higher after $M=20$ scoring rounds of \scorpio ($K=4$), on HMMT Feb 2025 (top) and LCBv6 (bottom). Fractions are averaged over questions and $20$ repetitions.}
  \label{fig:crossjudge_venn}
\end{figure}

Different judges can reach a similar level of ranking bias while making different mistakes. On LCBv6, Gemma-4-E4B and the self judge (Qwen3.5-4B) reach a similar squared bias at $M=20$ ($0.17$ and $0.18$; \autoref{fig:crossjudge}). \autoref{fig:crossjudge_venn} checks, for every pair of a correct and an incorrect candidate, whether the self judge and another judge each rank the correct candidate higher. Despite their similar bias, Gemma-4-E4B and the self judge disagree on $13.9\%$ of pairs: $6.2\%$ are ranked correctly only by the self judge and $7.7\%$ only by Gemma-4-E4B. Models with similar bias can therefore be biased on different pairs.

\FloatBarrier
\section{Experimental Details}
\label{app:exp-setup}

\subsection{Sampling Parameters and budgets}
\label{app:sampling}

We leverage SGLang \citep{zheng2024sglang} for LLM inference.

\textbf{Sampling parameters.} Following \citet{singh2026v_1}, we set the temperature for GPT-OSS-20B to $T=0.6$ for code generation and $T=1.0$ for mathematics and Reasoning Gym tasks, with top-$p$ sampling at $p=0.95$ throughout. For Qwen3.5-4B, we use the developer-recommended sampling parameters for instruct mode across all tasks: $T=1.0$, top-$p=0.95$, top-$k=20$, and a presence penalty of $1.5$.

\textbf{Budgets for the Judge.} When scoring candidates, all methods are provided a maximum
generation budget of $32{,}000$ tokens per LLM call.

Prompts for scoring responses are provided in \autoref{app:prompts}.

\subsection{Cost Calculation}
\label{app:cost-accounting}
In this section, we describe how the cost per question is calculated for each test-time scaling method discussed in the paper.

\textbf{Verification.}
The reported cost includes both generating the $N$ candidate solutions and verifying them through repeated evaluations. Let $\bar{g}_{\mathrm{in}}$ and $\bar{g}_{\mathrm{out}}$ denote the average input and output token counts per generation call, and let $\bar{v}_{\mathrm{in}}$ and $\bar{v}_{\mathrm{out}}$ denote the corresponding averages per verification call. For \scorpio with group size $K$, $M$ scoring rounds require $MN/K$ verification calls. The total cost per question is therefore
\[
\mathrm{Cost}
=
\frac{1}{10^6}
\left[
N\left(p_{\mathrm{in}}\bar{g}_{\mathrm{in}}
      +p_{\mathrm{out}}\bar{g}_{\mathrm{out}}\right)
+
\frac{MN}{K}
\left(p_{\mathrm{in}}\bar{v}_{\mathrm{in}}
      +p_{\mathrm{out}}\bar{v}_{\mathrm{out}}\right)
\right],
\]
where $p_{\mathrm{in}}$ and $p_{\mathrm{out}}$ are the input and output prices in dollars per million tokens. Verification input counts include the judging prompt, the problem, and the candidate solutions presented in each call. For other scoring methods, we use the actual number of verification calls in place of $MN/K$ and compute $\bar{v}_{\mathrm{in}}$ and $\bar{v}_{\mathrm{out}}$ from that method's input and output token counts.

\textbf{Search.}
For \scorpio and ThinkPRM, cost accounting follows the same principle as verification, but also includes evaluations of partial solutions during search. As described in \autoref{sec:tree_search}, trajectories are progressively pruned so that only a subset continues to completion. For a pruned trajectory, we count all generation and evaluation costs incurred before it is stopped. The total cost includes generation, intermediate evaluations, and verification of the completed solutions. For ThinkPRM, we use the same input and output token prices as the generating model. For DeepConf, judging intermediate states requires no additional model calls, so the cost includes only generation.

\textbf{Aggregation.}
The total cost includes generating the initial pool of $N$ solutions and the input and output token costs of every aggregation call across all rounds. When \scorpio is used between rounds, we also include the cost of scoring candidates for selection.

\subsection{Implementation Details of Verification Methods}
\textbf{V$_1$ }\citep{singh2026v_1}\textbf{.} We use the authors' \href{https://github.com/HarmanDotpy/pairwise-self-verification}{implementation}. Following their setup, for GPT-OSS-20B, the judge receives the final-channel output on math and Reasoning Gym tasks, while on LCBv6, the code block is extracted and provided as input. For Qwen3.5-4B (in instruct mode), which does not separate reasoning traces into distinct channels, we summarize the outputs using the \textit{same} model and provide these summaries to the judge. The additional summarization cost is \textit{included} in \textit{all} figures reported in the main and supplementary text.

\textbf{Pointwise scoring.} We use the implementation from the $\mathrm{V}_1$ codebase, replacing a single evaluation with repeated evaluations. Each response is scored multiple times by the judge, and its final score is the average of the resulting scores.

\textbf{LLM as a Verifier (PPT) }\citep{kwok2026llmasaverifiergeneralpurposeverificationframework}\textbf{.} For a fair comparison, we fix the evaluation criteria to be the same as those in our prompt ($C = 1$, as we use only a single evaluation criterion). We use the same rating scale described in their paper. For the experiments in the paper, the generating and evaluating models are the same.

\textbf{\scorpio.} Our implementation is based on the $\mathrm{V}_1$ codebase, with two differences: we replace $\mathrm{V}_1$'s Swiss Tournament strategy with a random grouping strategy, and we put $4$ elements in context together instead of two. Prompts for scoring complete responses are provided in \autoref{app:prompts-self-eval}.

\subsection{Implementation Details of Tree Search Methods}
\label{app:tree-search}

In this section, we describe the search algorithm used to obtain the results in \autoref{sec:tree_search}. We adapt the search algorithm of \citet{fu2025deepthinkconfidence} to perform value estimation: intermediate states are provided as inputs to a scorer, which reasons about them and outputs a score for each state. This reasoning incurs additional evaluation costs, but accurate scores can reduce generation costs by allowing unpromising trajectories to be pruned well before completion. We use the same sampling parameters for tree search as outlined in \autoref{app:sampling}.

\autoref{alg:tree-search} describes the procedure. Starting from $N$ trajectories, we alternate generation and state evaluation for $D=4$ stages, generating up to $L=2{,}048$ tokens per trajectory at each stage. Completed solutions are set aside, while the remaining partial solutions are scored and the lowest-scoring $25\%$ are pruned ($\rho=0.75$). After the final stage, surviving trajectories continue until completion or until the generation budget of $64{,}000$ tokens per trajectory is exhausted. The completed solutions are then scored to select the final answer. We now discuss the details specific to each method. We retain each baseline’s original final-answer selection strategy to preserve its strengths.

\textbf{\scorpio.} We use group size $K=4$ and $M=3$ scoring rounds for our main results. Intermediate state evaluations use the prompt in \autoref{app:prompts-partial}, while final verification uses the prompt in \autoref{app:prompts-self-eval}. We scale inference compute by increasing the initial pool size $N\in\{8,32,128\}$. 

\textbf{DeepConf} \citep{fu2025deepthinkconfidence}.  We use the authors' \href{https://github.com/facebookresearch/deepconf}{implementation} in the online setup, with the generation budget and sampling parameters outlined above. Test-time compute is scaled by increasing the number of candidates $N\in\{16,64,256\}$. These larger values of $N$ were chosen to roughly match the total costs of ThinkPRM and \scorpio. The number of warmup traces is $4$ for $N=16$ and $16$ for $N=64$ and $N=256$. We find that aggressive pruning ($\eta=10$) works better for Qwen3.5-4B, while conservative pruning ($\eta=90$) works better for GPT-OSS-20B, and we use these settings. We also report results for both values of $\eta$ in  \autoref{fig:app_deepconf_eta}. For final answer selection, we follow the paper and assign each trajectory a confidence score equal to the confidence of its lowest-confidence window. For the math benchmarks (AIME 2024, HMMT 2025), these scores are used as weights in a weighted majority vote. For LCBv6, we select the surviving trajectory with the highest  confidence.

\textbf{ThinkPRM} \citep{khalifa2025thinkprm}. Our search setup for ThinkPRM differs from the authors' original setup in two ways. First, their search uses double newlines to separate steps and a maximum tree depth of $20$, whereas ours uses a tree depth of $4$ with $2{,}048$  tokens generated per step. Second, we replace beam search with the early-stopping procedure adapted from DeepConf. These design choices were made to keep search-based strategies competitive in cost with verification-based methods, which generate complete solutions and then verify them. Building deeper trees with more frequent intermediate state evaluations, as in the original ThinkPRM setup, would incur substantially higher costs. We set $\rho=0.75$, retaining the top $75\%$ of partial solutions by score. For math tasks, we find that weighted majority voting works better than Best-of-$N$ selection. For coding tasks, we use Best-of-$N$ selection, as weighted majority voting is not possible.

\begin{algorithm}[t]
\caption{Tree Search with Intermediate Scoring}
\label{alg:tree-search}
\begin{algorithmic}[1]
\State \textbf{Inputs:} problem $x$; policy $\pi_\theta$; initial pool size $N$;
scorer $f$; group size $K$; scoring repeats $M$;
tree depth $D$; tokens per step $L$; retention fraction $\rho$.
\State $\gC \gets N$ empty trajectories
\State $\gC_{\mathrm{complete}} \gets \varnothing$
\For{$d \gets 1$ to $D$}
    \State Extend each trajectory in $\gC$ by at most $L$ tokens using $\pi_\theta$
    \State Move completed solutions from $\gC$ to $\gC_{\mathrm{complete}}$
    \If{$\gC=\varnothing$}
        \State \textbf{break}
    \EndIf
    \State Score partial solutions in $\gC$ using $f$ with $(K,M)$ and the prompt in \autoref{app:prompts-partial}
    \State $\gC \gets$ the top $\lceil \rho |\gC| \rceil$ partial solutions by score
\EndFor
\State Continue the remaining trajectories in $\gC$ until completion or the generation budget is exhausted
\State Add the resulting solutions to $\gC_{\mathrm{complete}}$
\State Score solutions in $\gC_{\mathrm{complete}}$ using the prompt in \autoref{app:prompts-self-eval}
\State \textbf{return} the highest-scoring solution
\end{algorithmic}
\end{algorithm}

\section{Additional Results}

\subsection{Selection Accuracy versus the pass@$N$ Gap}
\label{app:selection-passn-gap}

\begin{figure}[htbp]
  \centering
  \includegraphics[width=\textwidth]{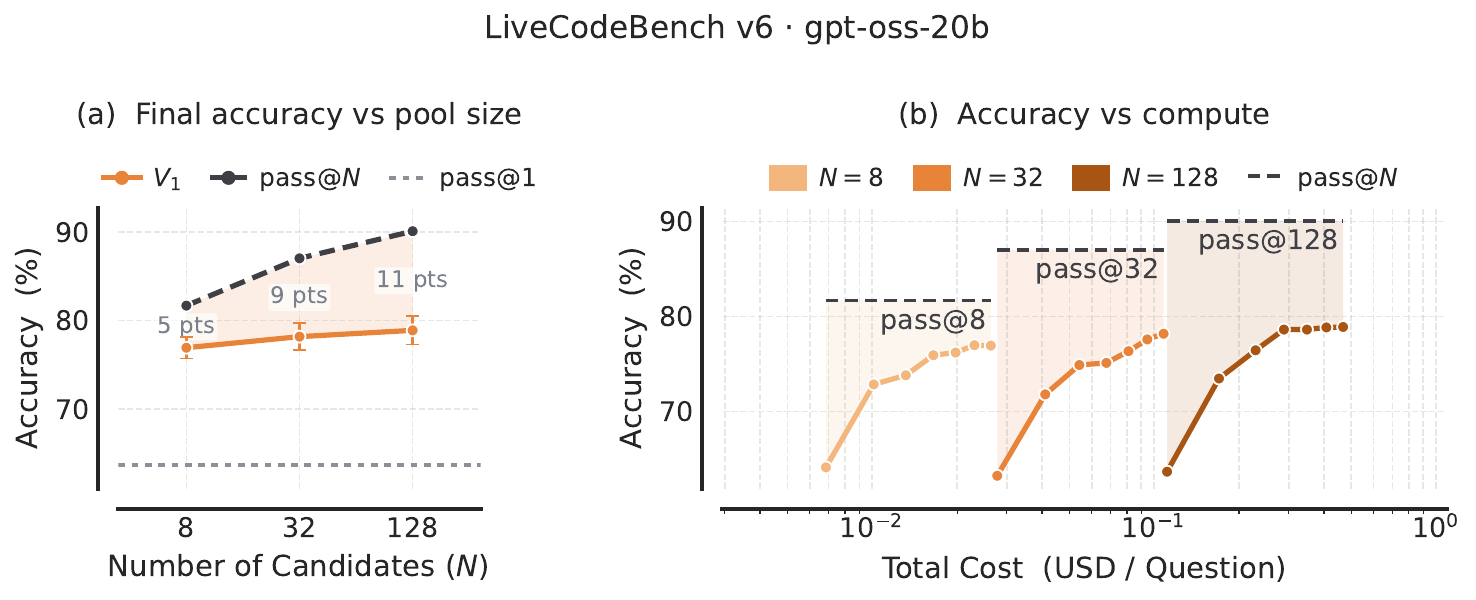}
  \caption{\textbf{\texorpdfstring{$\mathrm{V}_1$}{V1}\textnormal{~\citep{singh2026v_1}} accuracy plateaus well below \texorpdfstring{$\mathrm{pass@}N$}{pass@N}, and the gap widens as \texorpdfstring{$N$}{N} increases.} Results for GPT-OSS-20B on LCBv6. \textbf{(a)} Increasing the candidate pool from $N=8$ to $N=128$ yields modest gains in selection accuracy, while the gap to $\mathrm{pass@}N$ grows from roughly $5$ to $11$ percentage points. Points show means over $20$ seeds, with error bars indicating $95\%$ confidence intervals. \textbf{(b)} Accuracy plateaus as inference compute increases for each pool size, remaining below the corresponding $\mathrm{pass@}N$ ceiling (dashed lines). Inference compute is measured as the total cost of the input and output tokens of each LLM call; see \autoref{app:cost-accounting} for further details.} 
  \label{fig:app_v1_cost}
\end{figure}

\begin{figure}[htbp]
  \centering
  \includegraphics[width=\textwidth]{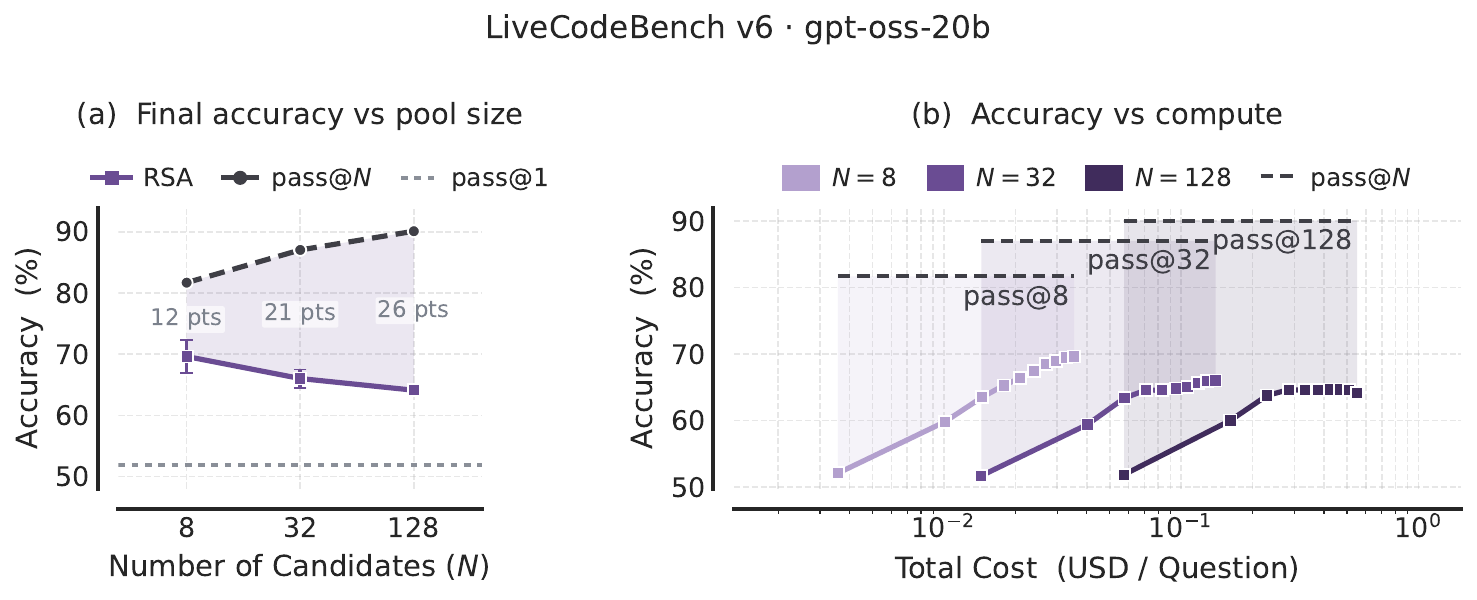}
  \caption{\textbf{RSA's ~\citep{venkatraman2025rsa} final accuracy decreases as the candidate pool grows.} Results for GPT-OSS-20B on LCBv6. \textbf{(a)} Increasing $N$ from $8$ to $128$ lowers final accuracy, widening the gap to $\mathrm{pass@}N$ from roughly $12$ to $26$ percentage points. Points show means over $8$ seeds, with error bars indicating $95\%$ confidence intervals. \textbf{(b)} Each curve tracks accuracy over successive aggregation rounds for a fixed $N$. Larger pools plateau at lower accuracy despite higher compute costs. Inference compute is measured as the total cost of the input and output tokens of each LLM call; see \autoref{app:cost-accounting} for further details.}
  \label{fig:app_rsa_cost}
\end{figure}

\begin{figure}[htbp]
  \centering
  \includegraphics[width=\textwidth]{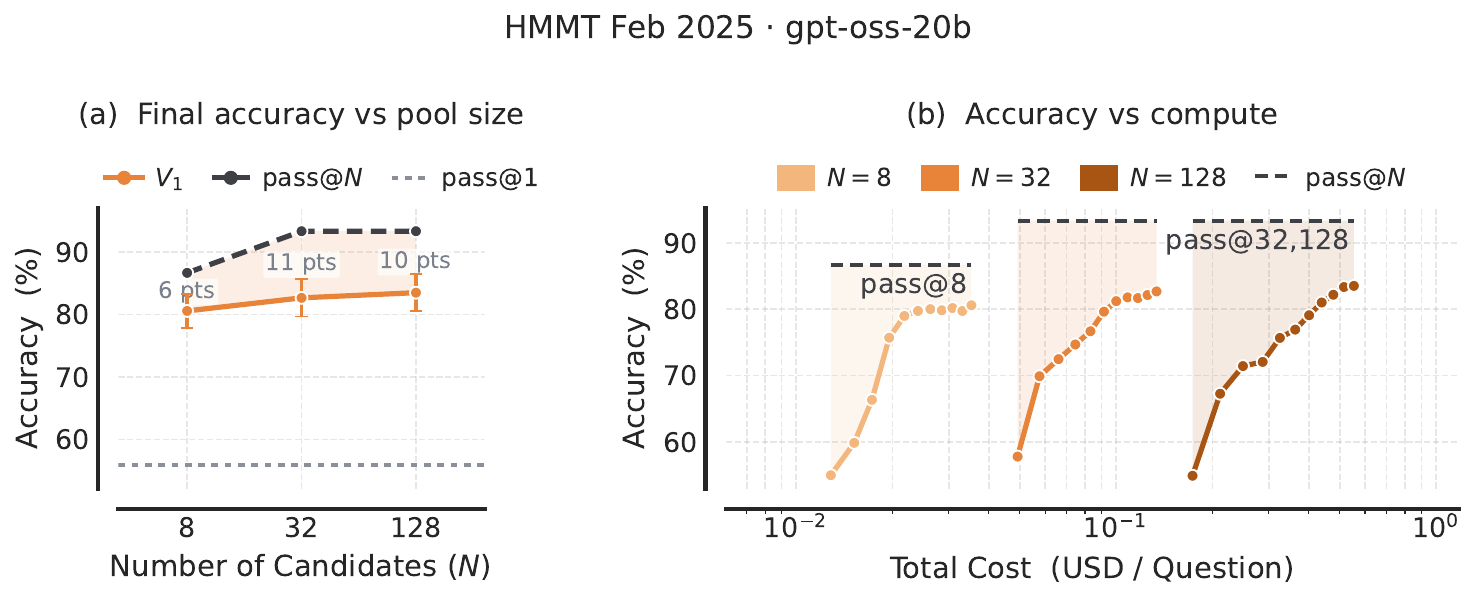}\caption{\textbf{$\mathrm{V}_1$\textnormal{~\citep{singh2026v_1}} also exhibits a widening selection gap on HMMT Feb 2025.} Results for GPT-OSS-20B. \textbf{(a)} Increasing $N$ from $8$ to $128$ yields modest gains in selection accuracy, while the gap to $\mathrm{pass@}N$ grows from roughly $6$ to $11$ percentage points. \textbf{(b)} Selection accuracy plateaus as inference compute increases, remaining below the corresponding $\mathrm{pass@}N$ ceiling (dashed lines). Inference compute is measured as the total cost of the input and output tokens of each LLM call; see \autoref{app:cost-accounting} for further details.}
  \label{fig:app_v1_cost_hmmt}
\end{figure}

In this section, we expand on the results in \autoref{fig:intro_b} and \autoref{fig:intro_c}, which report the final accuracies of $\mathrm{V}_1$~\citep{singh2026v_1} and RSA~\citep{venkatraman2025rsa} at different population sizes. To assess whether accuracy plateaus within the reported budget, we examine accuracy as a function of inference budget for GPT-OSS-20B on LCBv6. For $\mathrm{V}_1$, \autoref{fig:app_v1_cost} shows that accuracy plateaus for all values of $N$, with little improvement at higher budgets. Similar observations hold for RSA in \autoref{fig:app_rsa_cost}. We also report results on HMMT 2025 in \autoref{fig:app_v1_cost_hmmt} 
where a significant gap remains between the model's selection accuracy and oracle accuracy.


\subsection{Scaling Test-Time Compute on Verification}

In this section, we discuss additional results for scaling verification compute.

\begin{figure}[htbp]
  \centering
  \includegraphics[width=\textwidth]{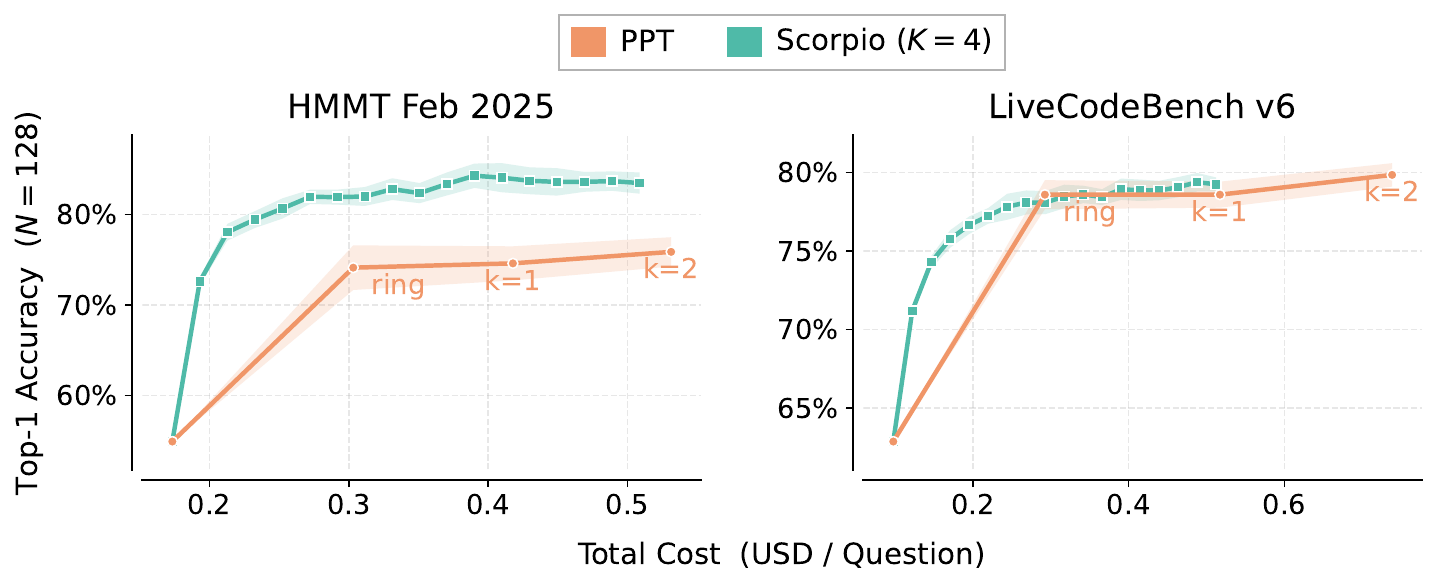}
  \caption{\textbf{\scorpio achieves higher accuracy on math and similar accuracy at lower cost on code compared with PPT \citep{kwok2026llmasaverifiergeneralpurposeverificationframework}.} Selection accuracy versus total inference cost for GPT-OSS-20B with $N=128$ candidates. On HMMT Feb 2025, \scorpio ($K=4$) improves peak accuracy by \textbf{8} percentage points over PPT. On LCBv6, it reaches PPT's accuracy at \textbf{1.5}$\times$ lower total cost. PPT's strategy proceeds in two phases: an initial ring phase ensures that every candidate is evaluated, followed by a tournament phase. Points labeled ``ring'' show accuracy after the first phase, while $k=1$ and $k=2$ indicate whether one or two evaluations are used per pair. Curves show mean accuracy, with shaded regions indicating $95\%$ confidence intervals. Inference compute is measured as the total dollar cost of all input and output tokens across LLM calls; see \autoref{app:cost-accounting} for details.}
  \label{fig:app_ppt_vs_scorpio}
\end{figure}

 \textbf{Comparison with PPT~\citep{kwok2026llmasaverifiergeneralpurposeverificationframework}.} In \autoref{fig:app_ppt_vs_scorpio}, we compare PPT with \scorpio ($K=4$) on HMMT Feb 2025 and LCBv6 using GPT-OSS-20B and $N=128$ candidates. On HMMT, \scorpio improves peak selection accuracy by \textbf{8} percentage points over PPT. On LCBv6, \scorpio reaches PPT's accuracy at \textbf{1.5$\times$} lower total cost.

\begin{figure}[htbp]
  \centering
  \includegraphics[width=\textwidth]{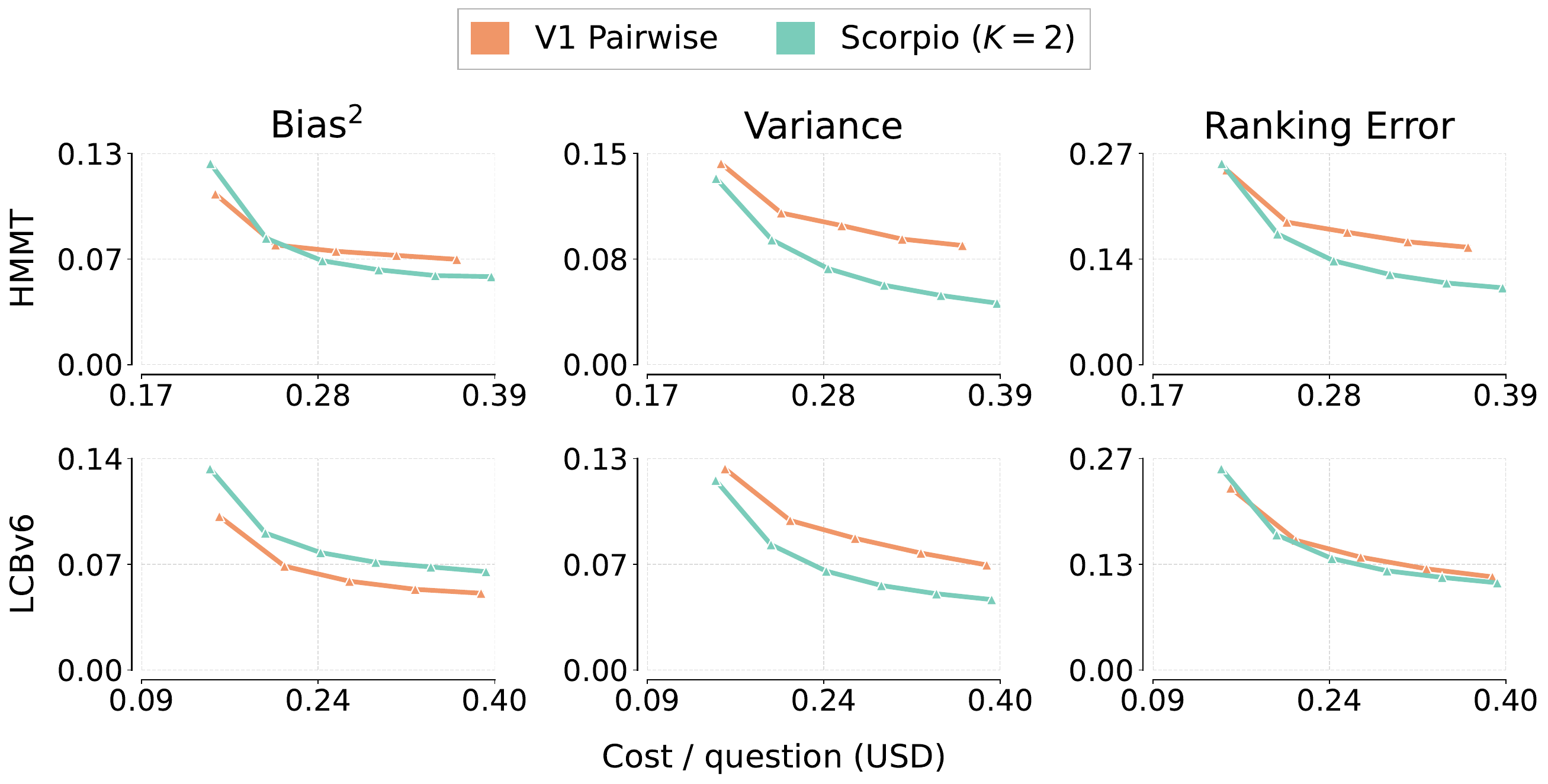}
  \caption{\textbf{Random pairing matches or outperforms tournament-based pairing.} Results for GPT-OSS-20B on HMMT and LCBv6. Bias-variance decomposition of ranking error for $\mathrm{V}_1$~\citep{singh2026v_1} and \scorpio ($K=2$) using GPT-OSS-20B. On both datasets, variance decreases quicker with \scorpio as compute increases. $\mathrm{V}_1$ attains lower bias on LCBv6, while \scorpio attains lower bias on HMMT. At higher budgets, \scorpio achieves similar or lower ranking error, demonstrating that a simpler pairing strategy can be equally or more effective than tournament based strategies. Inference compute is measured as the total dollar cost of all input and output tokens across LLM calls; see \autoref{app:cost-accounting} for details.}
  \label{fig:app_v1_vs_k2_biasvar}
\end{figure}

 \textbf{Comparing V$_{1}$'s pairing strategy with random pairing strategy.}  To isolate the effect of pairing strategy from group size, \autoref{fig:app_v1_vs_k2_biasvar} compares $\mathrm{V}_1$ with \scorpio at $K=2$ using GPT-OSS-20B. Both methods evaluate two candidates per call, but $\mathrm{V}_1$ pairs candidates based on score similarity, whereas \scorpio pairs them randomly. On both HMMT and LCBv6, random pairing reduces variance faster as compute increases.

\subsection{Ranking Quality for Qwen3.5-4B (Instruct Mode)}
\label{app:qwen}

In this section, we present results for the bias-variance decomposition for Qwen3.5-4B.

\begin{figure}[htbp]
  \centering
  \includegraphics[width=\textwidth]{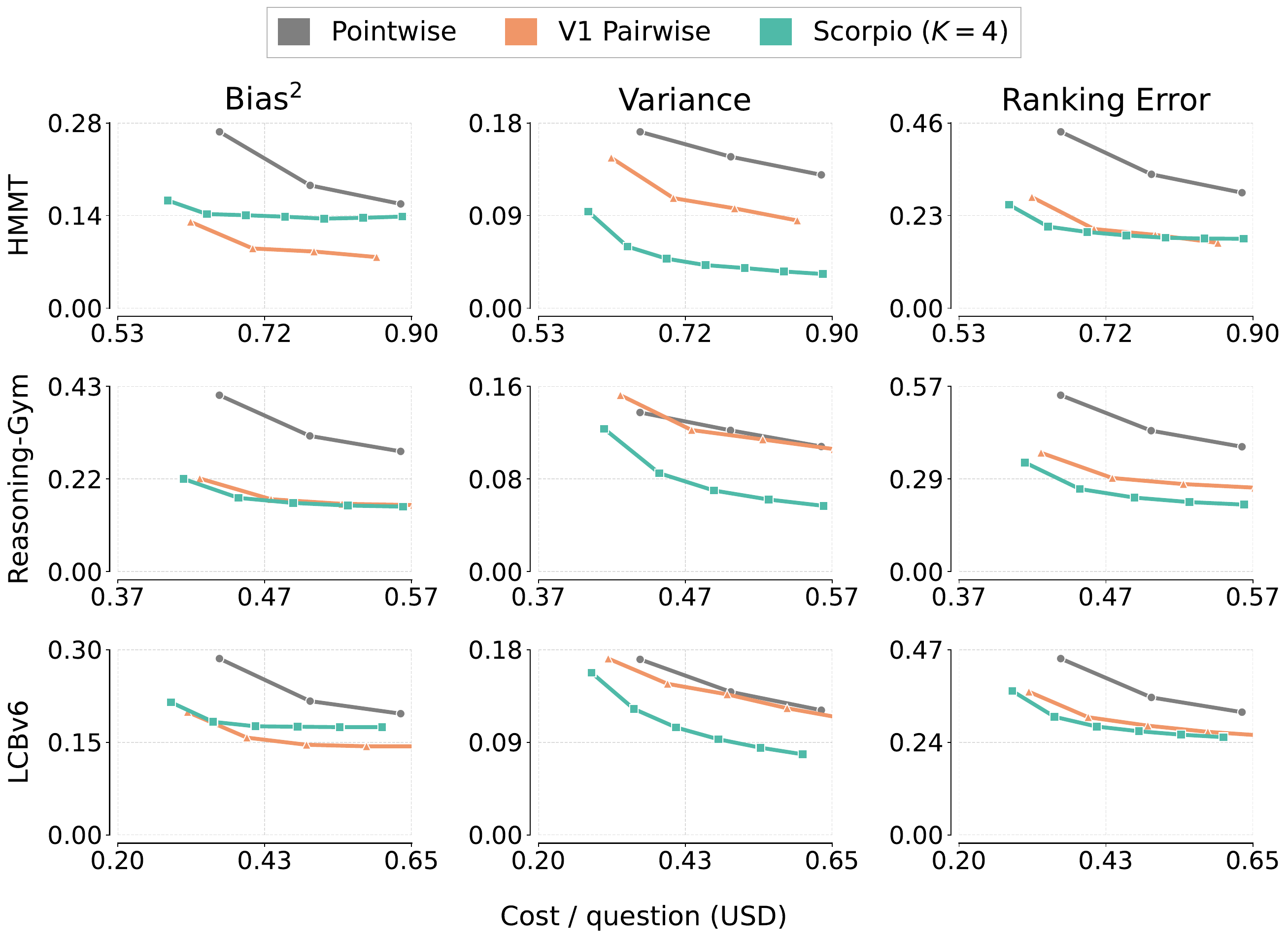}
  \caption{\textbf{Decomposition of the ranking error as a bias term and a variance term.} Results for Qwen3.5-4B on HMMT, Reasoning-Gym and LCBv6. While the variance term generally vanishes with increased compute, the bias term converges to a plateau (see: ~\autoref{app:bias-dec}). \scorpio's groupwise scoring leads to a higher variance reduction rate than pointwise or V1's \citep{singh2026v_1} pairwise scoring. Compute is measured as the total dollar cost of all input and output tokens across LLM calls; see \autoref{app:cost-accounting} for further details.}
  \label{fig:qwen_biasvar}
\end{figure}

\textbf{Bias-variance decomposition.} \autoref{fig:qwen_biasvar} shows the bias-variance decomposition of ranking error for Qwen3.5-4B. Joint evaluation reduces bias compared with pointwise scoring. While $\mathrm{V}_1$ attains lower bias than \scorpio on HMMT and LCBv6 at higher budgets, \scorpio reduces variance faster, achieving similar or lower ranking error.

\begin{figure}[htbp]
  \centering
  \includegraphics[width=\textwidth]{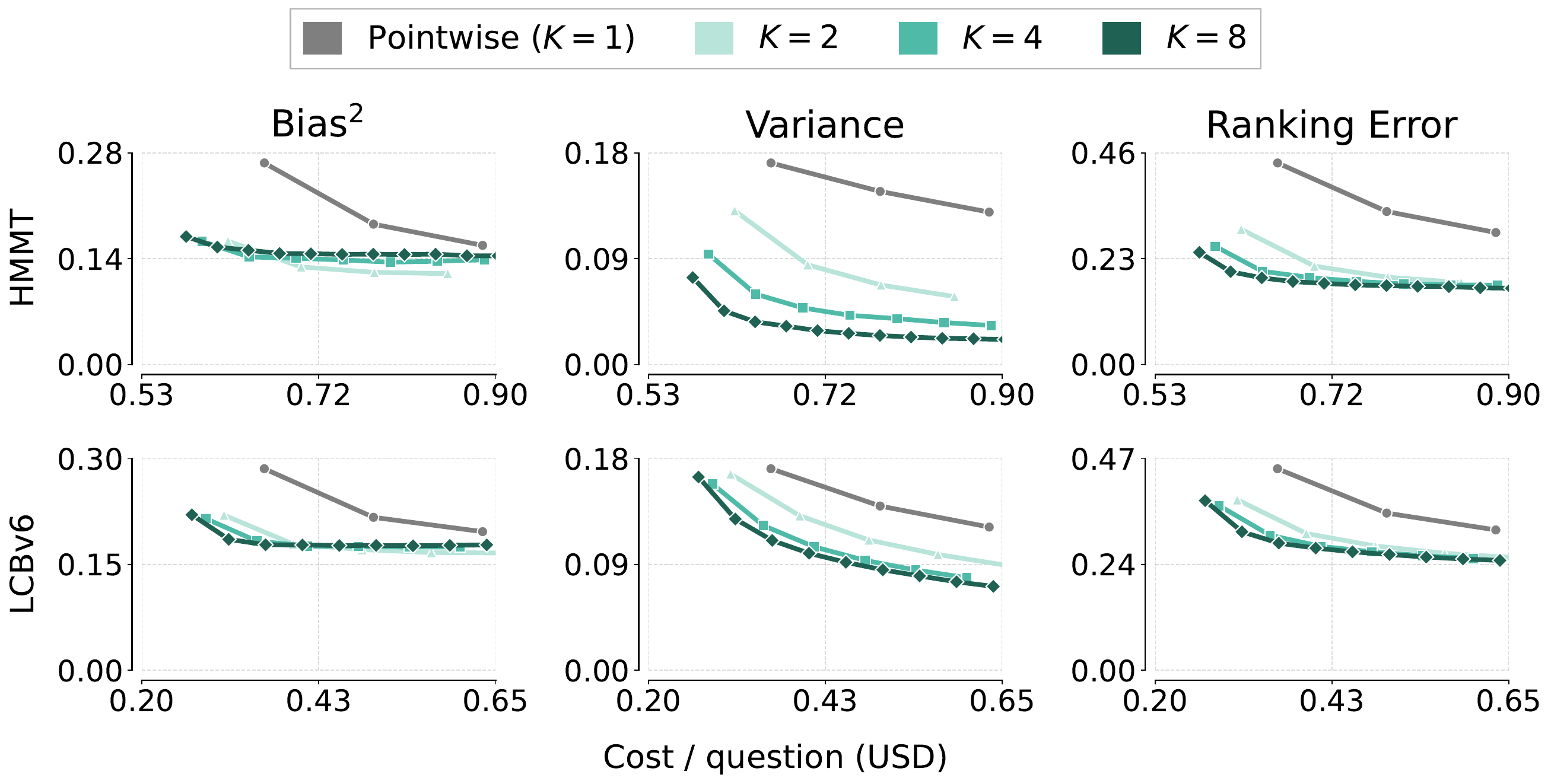}
  \caption{\textbf{Larger groups reduce variance faster but can increase bias.} Results for Qwen3.5-4B on HMMT and LCBv6. Increasing group size allows more evaluations within the same cost budget, reducing variance faster. However, these gains can come at the expense of bias: on LCBv6, $K=8$ yields higher bias than smaller groups. Compute is measured as the total dollar cost of all input and output tokens across LLM calls; see \autoref{app:cost-accounting} for further details.}
  \label{fig:kladder_qwen}
\end{figure}

\textbf{Larger groups help with variance reduction but can hurt bias.} \autoref{fig:kladder_qwen} shows how group size affects bias and variance for Qwen3.5-4B. Larger groups allow more evaluations within the same cost budget, helping reduce variance. However, increasing group size does not always reduce bias, highlighting the tradeoff between obtaining more score estimates and comparing more candidates in a single context.

\subsection{Scaling Test Time Compute Using Tree-Search}
\label{app:eta-threshold}

In this section, we present additional results on scaling the compute spent on test-time search. 

\begin{figure}[htbp]
  \centering
  \includegraphics[width=0.663\textwidth]{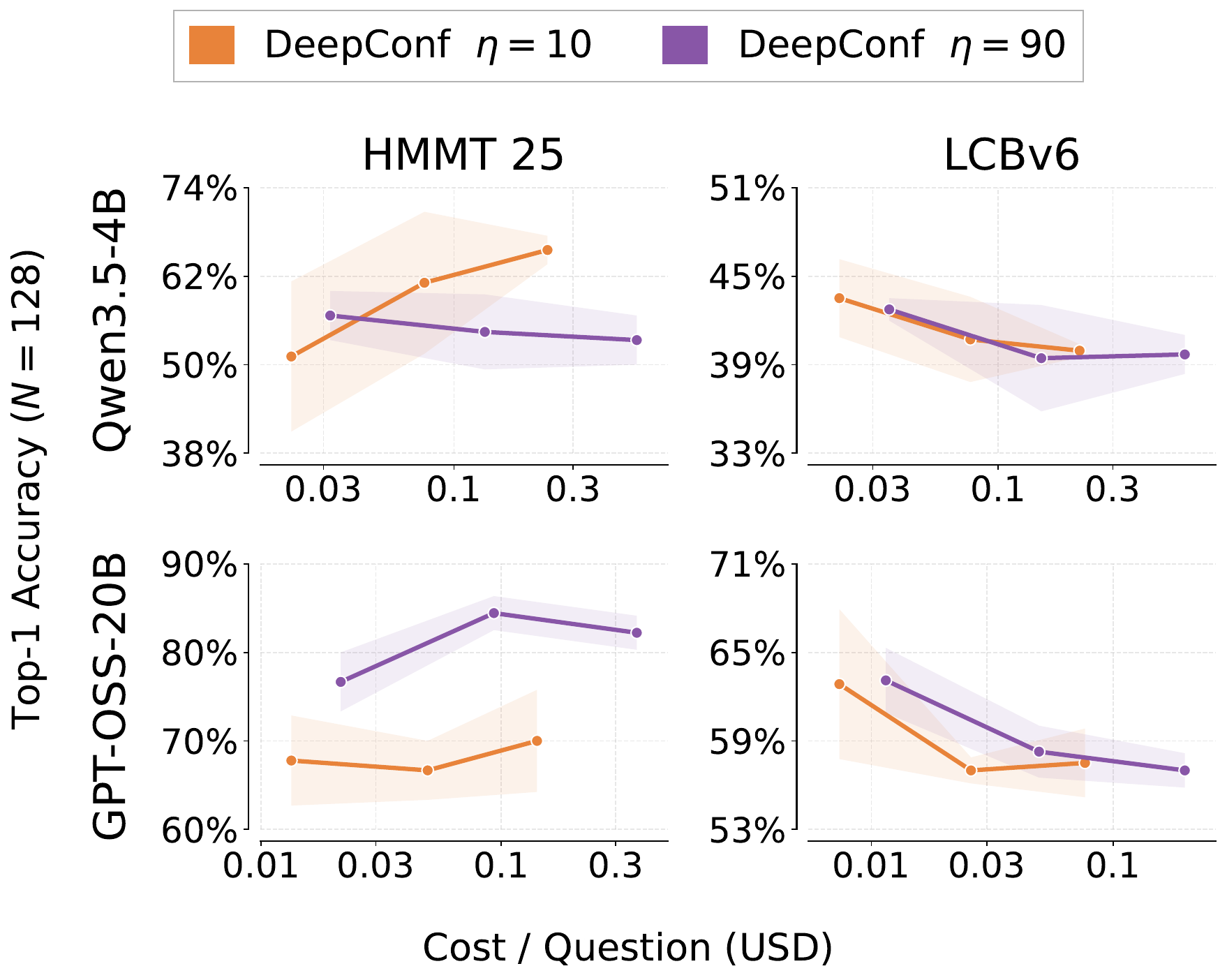}
  \caption{\textbf{DeepConf's \citep{fu2025deepthinkconfidence} pruning threshold depends on the model.} Top-1 Accuracy versus total inference cost for DeepConf with aggressive ($\eta=10$) and conservative ($\eta=90$) pruning. Aggressive pruning generally works better for Qwen3.5-4B, while conservative pruning works better for GPT-OSS-20B. Compute is measured as the total dollar cost of all input and output tokens across LLM calls; see \autoref{app:cost-accounting} for further details.}
  \label{fig:app_deepconf_eta}
\end{figure}

\textbf{DeepConf's~\citep{fu2025deepthinkconfidence} pruning threshold depends on the model.} \autoref{fig:app_deepconf_eta} compares aggressive ($\eta=10$) and conservative ($\eta=90$) pruning on HMMT Feb 2025 and LCBv6. Aggressive pruning generally works better for Qwen3.5-4B, while conservative pruning works better for GPT-OSS-20B. We therefore use $\eta=10$ for Qwen3.5-4B and $\eta=90$ for GPT-OSS-20B in our results.

\begin{figure}[htbp]
  \centering
  \includegraphics[width=0.85\textwidth]{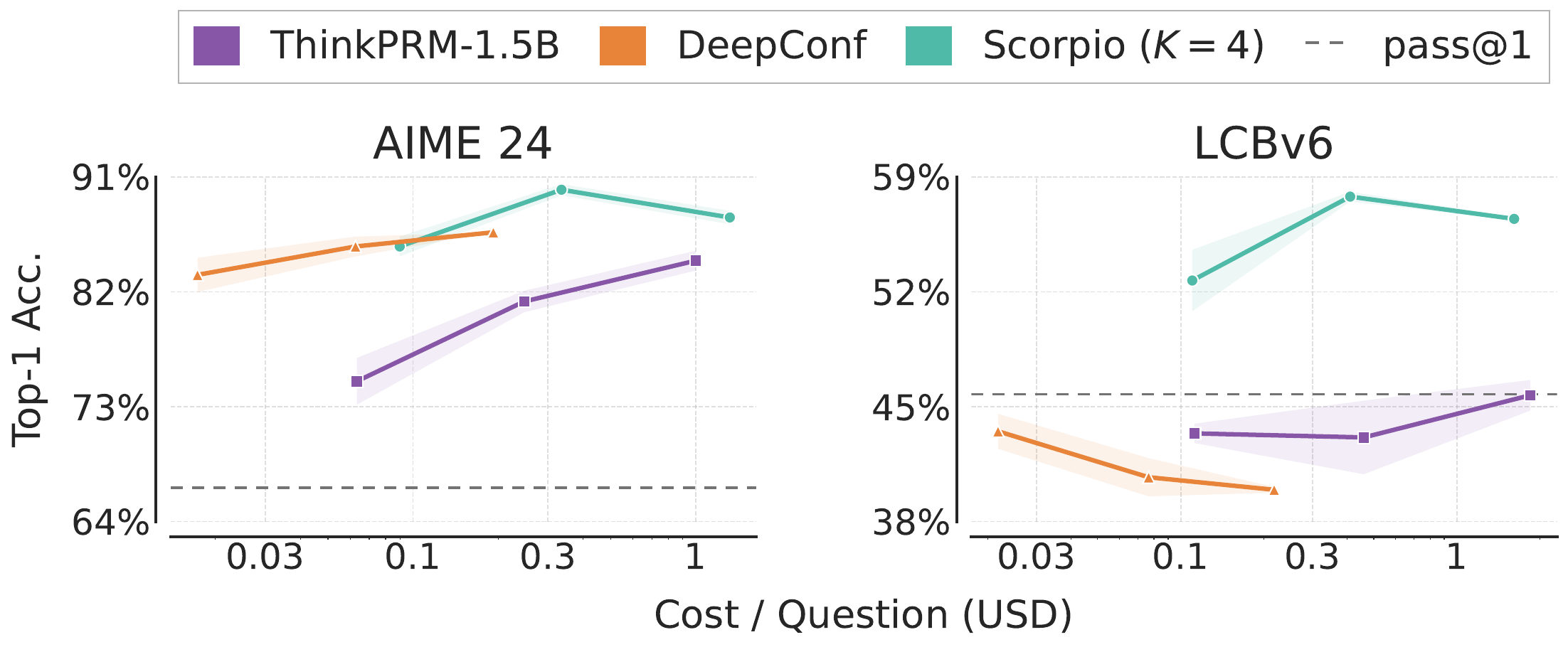}
  \caption{\textbf{Test-time search using \scorpio matches or outperforms prior methods.} Results for Qwen3.5-4B on AIME 2024 and LCBv6. Top-1 accuracy of test-time search versus total inference cost on AIME 2024 and LCBv6 for Qwen3.5-4B. We vary the number of solutions we start with: $N\in\{8,32,128\}$ for \scorpio and ThinkPRM, and $N\in\{16,64,256\}$ for DeepConf to keep total costs comparable. Inference compute is measured as the total dollar cost of all input and output tokens across LLM calls; see \autoref{app:cost-accounting} for details.}
  \label{fig:value_nscaling_qwen}
\end{figure}

\textbf{\scorpio improves test-time search on Qwen3.5-4B.} \autoref{fig:value_nscaling_qwen} compares \scorpio, ThinkPRM, and DeepConf on AIME 2024 and LCBv6. On AIME 2024, DeepConf and \scorpio attain similar peak accuracies, with \scorpio exceeding ThinkPRM by $6$ percentage points. On LCBv6, \scorpio improves peak accuracy over the strongest baseline by $12$ percentage points.

\begin{figure}[htbp]
  \centering
  \includegraphics[width=0.95\textwidth]{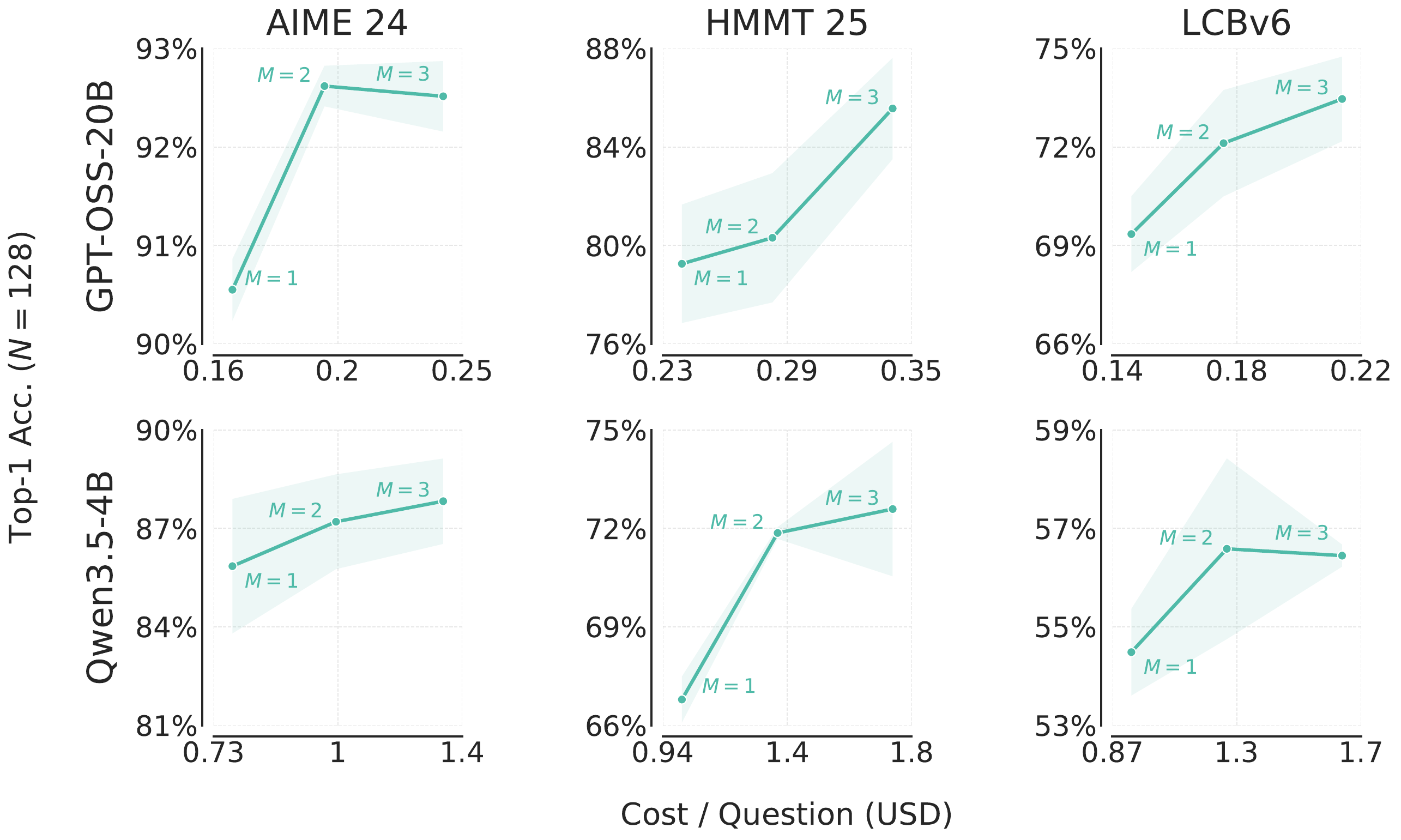}
  \caption{\textbf{Repeated evaluation of partial solutions improves search accuracy.} Top-1 accuracy of test-time search using \scorpio versus total inference cost on AIME 2024, HMMT Feb 2025, and LCBv6. We fix the number of solutions we start with at $N=128$ and vary the number of times the random grouping and scoring process is repeated, $M\in\{1,2,3\}$. Repeated evaluations improve accuracy over $M=1$ across all three datasets. Inference compute is measured as the total dollar cost of all input and output tokens across LLM calls; see \autoref{app:cost-accounting} for details.}
  \label{fig:m_scaling}
\end{figure}

\textbf{Repeated evaluation of partial solutions improves search accuracy.} In \autoref{fig:m_scaling}, we fix the number of solutions we start with at $N=128$ and vary the number of times the random grouping and scoring process in \autoref{alg:tree-search} is repeated, $M\in\{1,2,3\}$. Across both models and all three datasets, increasing $M$ from $1$ to $3$ improves accuracy by $2$ to $6$ percentage points, showing that spending additional compute on state evaluation improves the accuracy of test-time search.

\subsection{Scaling Test-Time Compute Using Aggregation}
\label{app:agg-scaling}

In this section, we examine how candidate pool size affects the amplification of tool-call responses in RSA and how \scorpio limits this behavior.

\begin{figure}[htbp]
  \centering
  \includegraphics[width=\textwidth]{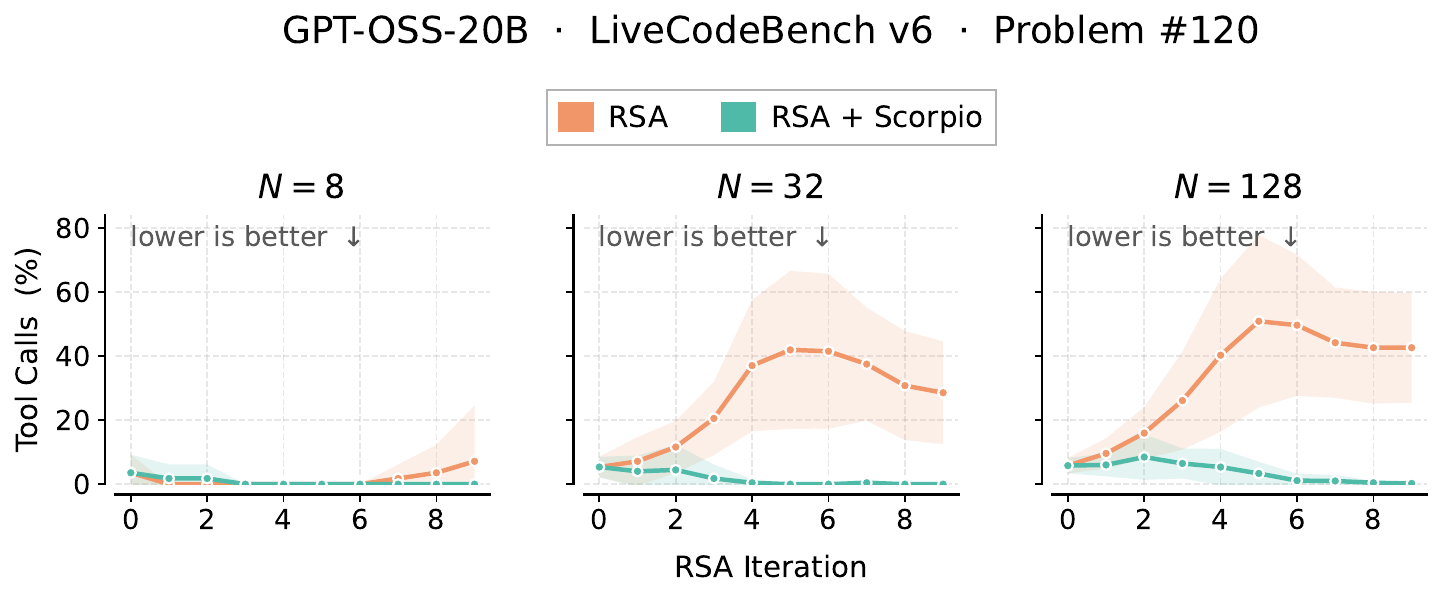}
  \caption{\textbf{Larger pools amplify tool-call responses in RSA, while \scorpio limits their amplification.} Proportion of responses containing tool calls across aggregation rounds for a single LCBv6 problem using GPT-OSS-20B. This is a single-turn task that requires a solution in a Python code block. Some responses attempt tool calls to test their logic, which we identify using the pattern \texttt{to=[A-Za-z\_]}. The harness does not support tool calls, and $95\%$ of these responses receive zero reward. Tool-call responses remain uncommon with $N=8$ but become increasingly common with $N=32$ and $N=128$ as aggregation rounds progress. In RSA + \scorpio, candidates are scored using \scorpio after each round, and only the top half of candidates are used to build aggregation sets for the next round. This selection reduces the proportion of tool-call responses to nearly zero across all three pool sizes.}
  \label{fig:app_rsa_toolcalls}
\end{figure}

\textbf{Larger pools could potentially amplify tool-call responses in RSA, while \scorpio can limit their amplification.}
\autoref{fig:app_rsa_toolcalls} shows that tool-call responses remain uncommon with $N=8$ but become increasingly common across RSA rounds with $N=32$ and $N=128$. In RSA + \scorpio, scoring candidates after each round and retaining the top half for subsequent aggregation reduces the proportion of these responses to nearly zero across all three pool sizes.

\clearpage
\section{Scoring Candidates Using LLMs}
\label{app:prompts}

\subsection{Prompt used for \scorpio scoring of complete candidates}
\label{app:prompts-self-eval}

Our scoring prompt is largely based on the pairwise scoring prompt of \citet{singh2026v_1},
with the change that we provide $K$ solutions instead of two and ask the model to output a
score for each solution.

\begin{promptbox}{Prompt for coding tasks (LiveCodeBench)}
\small
\fontfamily{cmtt}\selectfont
You are an expert code reviewer. Compare K solutions to a programming problem and rate their correctness.\\[0.5em]
\textbf{**Evaluation Guidelines:**}\\
- Analyze the problem's requirements and constraints.\\
- Mentally trace each solution with test cases (including edge cases) to verify correctness.\\\\
- If multiple solutions appear correct, prefer the more robust and fault-tolerant ones.\\[0.5em]\\
\textbf{**Problem**}\\
\{problem\}\\[0.5em]
\textbf{**Solution 1**}\\
\{solution 1\}\\
   ...\\
\textbf{**Solution K**}\\
\{solution K\}\\[0.5em]
\textbf{**Output Format:**} First, compare and contrast the solutions provided in context. Then provide your step-by-step reasoning following the evaluation guidelines above. Then, on separate new lines, provide your final ratings using the EXACT tags below.\\
<rating\_1>INTEGER\_1\_TO\_10</rating\_1>\\
   ...\\
<rating\_K>INTEGER\_1\_TO\_10</rating\_K>\\[0.5em]
\textbf{**Rating Rules:**}\\
- Rate correctness on a 1-10 scale (10 = correct \& robust,\\
  5 = borderline, 1 = incorrect).\\
- Higher rating wins. Equal ratings imply ties. Please provide your analysis now.
\end{promptbox}

\clearpage
\begin{promptbox}{Prompt for math (HMMT 25)}
\small
\fontfamily{cmtt}\selectfont
You are an expert math contest grader. Compare K submissions and rate correctness based solely on the final answer.\\[0.5em]
\textbf{**Evaluation Guidelines:**}\\
- Extract each submission's final answer. Use any provided reasoning only to help you assess whether the stated final answer is trustworthy. Do not award credit for method quality or rigor.\\\\
- Carefully analyze the problem statement and the submissions to assess whether each final answer is correct. Grade only the final answer.\\[0.5em]\\
\textbf{**Problem**}\\
\{problem\}\\[0.5em]
\textbf{**Solution 1**}\\
\{solution 1\}\\
   ...\\
\textbf{**Solution K**}\\
\{solution K\}\\[0.5em]
\textbf{**Output Format:**} First, compare and contrast the solutions provided in context. Then provide your reasoning (what checks you performed) following the evaluation guidelines above. Then, on separate new lines, give ratings using the EXACT tags below.\\
<rating\_1>INTEGER\_1\_TO\_10</rating\_1>\\
   ...\\
<rating\_K>INTEGER\_1\_TO\_10</rating\_K>\\[0.5em]
\textbf{**Rating Rules:**}\\
- Rate correctness on a 1-10 scale (10 = certainly correct, 8 = very likely correct, 5 = uncertain/borderline, 3 = likely incorrect,\\\\
  1 = certainly incorrect).\\
- Higher rating wins. Equal ratings imply ties. Please provide your analysis now.
\end{promptbox}

For reasoning-gym games we use the math prompt above verbatim, replacing \texttt{expert math contest grader} with \texttt{expert grader for a reasoning-puzzle benchmark}.

\subsection{Prompt used for \scorpio scoring of partial candidates}
\label{app:prompts-partial}

For the tree-search experiments in \autoref{sec:tree_search}, \scorpio scores \emph{in-progress} trajectories rather than completed solutions. The prompt therefore asks for the probability that each candidate will reach a correct final answer within the remaining token budget if generation continues. A single template is used for both the math (HMMT Feb 2025) and coding (LCBv6) benchmarks. The judge is shown each candidate's reasoning so far, the number of tokens consumed, and the remaining token budget.

\clearpage
\begin{promptbox}{Prompt for scoring partial candidates (math and code)}
\small
\fontfamily{cmtt}\selectfont
You are evaluating K in-progress attempts at the SAME problem, labelled Candidate 1 to Candidate K. For EACH candidate, estimate the probability that it reaches a correct final answer when continued. Do not solve the problem yourself.\\[0.5em]
\#\# Generation budget\\
- Each attempt has used \{tokens used\} tokens so far\\
- Each attempt can use at most \{remaining budget\} more tokens to continue\\[0.5em] \#\# Problem\\\\
\{problem\}\\[0.5em]
\#\# Candidates \#\#\# Candidate 1\\\\
\{partial solution 1\}\\
   ... \#\#\# Candidate K\\\\
\{partial solution K\}\\[0.5em]
\#\# How to judge Do this in two stages, in a single response.\\[0.5em]
Stage 1 \textemdash{} compare and contrast. Looking at all K attempts together, compare and contrast the approaches they take. For each candidate, work through its reasoning and CHECK it \textemdash{} verify the key claims and steps (calculations, logic, edge cases \textemdash{} whatever the task involves) \textemdash{} and identify what is correct and what is incorrect or questionable about it. Where the candidates agree or diverge is itself evidence about what is right.\\[0.5em]
Stage 2 \textemdash{} score each candidate. Using that analysis, assign each candidate a score = the probability that it reaches a correct final answer when continued. For each candidate, weigh:\\
- Correctness so far \textemdash{} are the written steps right, and are any mistakes minor (recoverable) or fatal (cannot reach the answer without changing approach)?\\\\
- Approach \textemdash{} is it a valid method that, if continued correctly, can reach the answer?\\\\
- Budget feasibility \textemdash{} with \{remaining budget\} tokens left, can it realistically finish?\\ Incompleteness by itself is not a penalty.\\[0.5em] \#\# Score meaning (per candidate) Score = probability that THAT candidate reaches a correct final answer when continued, on a 1-10 scale:\\\\
10 = almost certain   7 = likely   5 = about 50\%\\
 3 = unlikely         1 = almost no chance\\[0.5em]
\#\# Output format First write your Stage 1 compare-and-contrast analysis across all candidates. Then go through the candidates IN ORDER \textemdash{} for each, reason about its state and then give its score:\\[0.5em]
Candidate 1: <reasoning about this candidate's state>\\
Score: N\\
Candidate 2: <reasoning about this candidate's state>\\
Score: N\\
... continue through Candidate K\\[0.5em]
Then end with EXACTLY one final line repeating all K scores in order, comma-separated:\\[0.5em]
<<<SCORES: N1, N2, ..., NK>>>\\[0.5em]
Score every candidate (an integer from 1 to 10).
\end{promptbox}

\end{document}